%% file: main.tex
\documentclass[10pt,twocolumn,letterpaper]{article}

\usepackage[pagenumbers]{cvpr} 


\definecolor{cvprblue}{rgb}{0.21,0.49,0.74}
\usepackage[pagebackref,breaklinks,colorlinks,allcolors=cvprblue]{hyperref}
\usepackage{multirow}

\def\paperID{7676} 
\def\confName{CVPR}
\def\confYear{2025}

\title{MedUnifier: Unifying Vision-and-Language Pre-training on Medical Data with Vision
Generation Task using Discrete Visual Representations}

\author{Ziyang Zhang$^{1, 2}$, Yang Yu$^3$, Yucheng Chen$^{1, 4}$, Xulei Yang$^{3, *}$, Si Yong Yeo$^{1, 4, *}$\\
$^1$MedVisAI Lab, $^2$ECE, Northwestern University, $^3$Institute for Infocomm Research (I$^{2}$R),\\ A*STAR, Singapore, $^4$Lee Kong Chian School of Medicine, Nanyang Technological University\\
{\tt\small $*$ Corresponding authors}
}

\begin{document}
\maketitle
\input{sec/0_abstract}    
\input{sec/1_intro}
\input{sec/2_related}
\input{sec/3_method}
\input{sec/4_experiments}

\input{sec/5_conclusion}
\input{sec/6_acknowledgement}
{
    \small
    \bibliographystyle{ieeenat_fullname}
    \bibliography{main}
}

\input{sec/X_suppl}

\end{document}

%% file: sec/0_abstract.tex
\begin{abstract}
Despite significant progress in Vision-Language Pre-training (VLP), current approaches predominantly emphasize feature extraction and cross-modal comprehension, with limited attention to generating or transforming visual content. This gap hinders the model's ability to synthesize coherent and novel visual representations from textual prompts, thereby reducing the effectiveness of multi-modal learning. In this work, we propose \textbf{MedUnifier}, a unified VLP framework tailored for medical data. MedUnifier seamlessly integrates text-grounded image generation capabilities with multi-modal learning strategies, including image-text contrastive alignment, image-text matching and image-grounded text generation. Unlike traditional methods that reply on continuous visual representations, our approach employs visual vector quantization, which not only facilitates a more cohesive learning strategy for cross-modal understanding but also enhances multi-modal generation quality by effectively leveraging discrete representations. Our framework's effectiveness is evidenced by the experiments on established benchmarks, including uni-modal tasks, cross-modal tasks, and multi-modal tasks, where it achieves state-of-the-art performance across various tasks. MedUnifier also offers a highly adaptable tool for a wide range of language and vision tasks in healthcare, marking advancement toward the development of a generalizable AI model for medical applications.
\end{abstract}


%% file: sec/1_intro.tex
\section{Introduction}
\label{sec:intro}



The rapid growth of medical imaging datasets has accelerated the development of deep-learning models to enhance clinical decision-making processes. However, annotating these extensive datasets requires specialized expertise, making large-scale annotation unfeasible. To overcome this limitation, one effective approach is to leverage associated medical reports containing detailed diagnostic descriptions provided by radiologists \cite{reyes2020interpretability}. In recent years, deep learning models that utilize multi-modal data as inputs have drawn more attention, driven by advancements in attention mechanisms or transformer-based models \cite{li2023lvit, zhong2023ariadne, huh2023improving}. 
 
Accordingly, vision-and-language pre-training (VLP) models have been developed, many drawing inspiration from the foundational CLIP model \cite{radford2021learning}. These models primarily leverage a dual-encoder approach, consisting of an image encoder and a text encoder, to independently extract uni-modal features. They aim to maximize cosine similarity between paired data via contrastive learning. Researchers have further enhanced these models by incorporating domain-specific knowledge and making targeted adjustments to the original CLIP, resulting in label-efficient adaptations \cite{zhang2022contrastive, huang2021gloria, zhou2022generalized, you2023cxr, wu2023medklip, wang2022medclip}. In addition, fusion-based encoders have attracted considerable attention. These fusion models utilize self-attention or co-attention mechanisms to achieve early integration of visual and textual modalities \cite{li2021align, bao2022vlmo, li2022blip}. This joint processing enables the learning of multi-modal representations that are crucial for tasks requiring complex multi-modal reasoning, such as medical visual question answering. For fusion models, cross-modal matching with hard sampling strategies is employed to strengthen correlations between matched data. Image-grounded text understanding with masked language modelling (MLM), originally developed for the BERT \cite{devlin2018bert}, is also deployed to enhance multi-model interaction. PTUnifier \cite{chen2023towards}, as an example, focused on multi-modal understanding yet did not possess generative ability during pre-training, thus necessitating the use of an additional language decoder and fine-tuning for language generative tasks. Meanwhile, image-grounded text generation with the causal language modelling (CLM) is often applied to facilitate vision-grounded language generation tasks. However, we observe that current VLP approaches often overlook the generation of visual content, limiting the model's capacity to produce coherent and novel visual representations based on textual or multi-modal prompts, thus reducing the potential of multi-modal learning. Although recent studies have integrated masked image modelling (MIM) into the VLP framework \cite{zhou2022advancing, chen2023contrastive}, this approach does not fully enable the generation of comprehensive visual content nor capture detailed visual information effectively.



\begin{figure}[tp]
    \centering{\includegraphics[width=\linewidth]{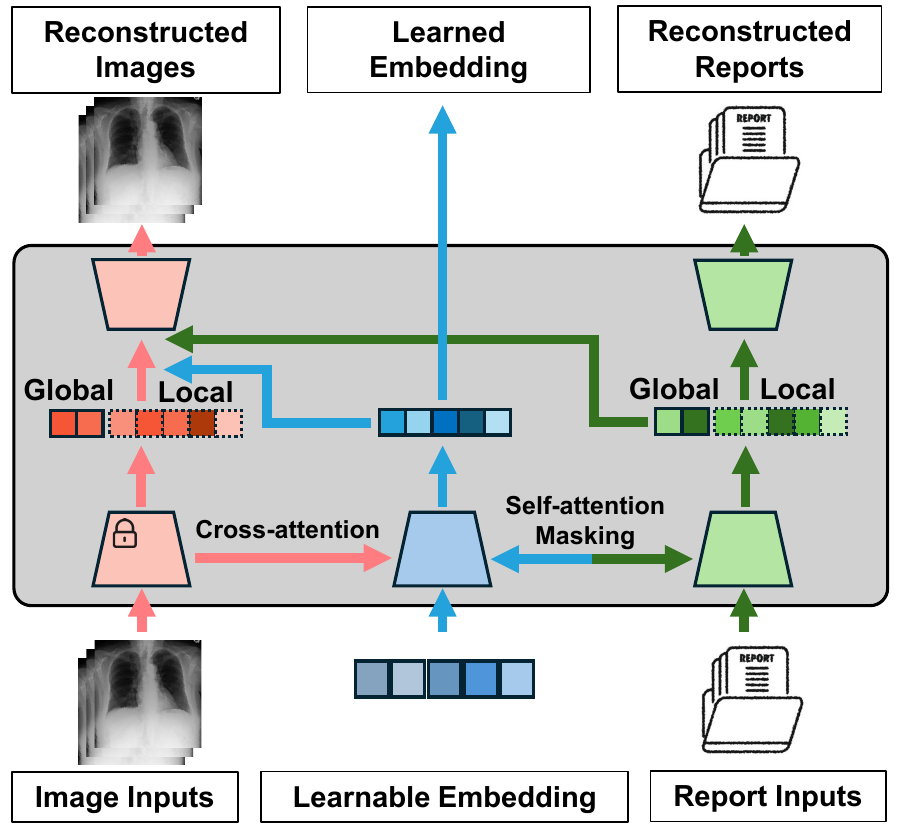}}
    \caption{Our MedUnifier framework incorporates learnable embeddings to enable multi-modal interactions. The red components focus on the initial extraction of visual features and the reconstruction of medical images. The green elements are dedicated to the modelling and interpretation of medical reports. Meanwhile, the blue components apply a range of attention-masking strategies to achieve a comprehensive fusion of image and text representations.}
    \label{fig1}
\end{figure}

 
In this paper, we present \textbf{MedUnifier}, a unified VLP framework for medical data (Figure \ref{fig1}), designed to seamlessly integrate text-grounded image generation with advanced multi-modal learning strategies, including image-text contrastive alignment, image-text matching, and image-grounded text generation. Our approach begins with a Transformer model featuring learnable embeddings, inspired by BLIP-2 \cite{li2023blip}, a 12-layer transformer encoder as the trainable component, paired with a pre-trained visual encoder that embeds preliminary visual features. To extend its functionality for vision generation, we introduce a novel learning objective, termed text-grounded image generation (\textbf{TIG}) loss. This objective leverages vector-quantization to facilitate discrete visual representation learning \cite{van2017neural}, guiding vision generation using textual data. Additionally, we devise a novel latent adapter to connect the base model with the image generation module, enabling end-to-end co-training with three other learning objectives: image-text contrastive (\textbf{ITC}), image-text matching (\textbf{ITM}), and image-grounded text generation (\textbf{ITG}) losses. To our knowledge, \textbf{MedUnifier} is the first model to adapt learnable embeddings to the medical domain, bridging the gap between existing VLP paradigms and text-grounded image generation to enhance multi-modal alignment. The main contribution of this study is listed below:
 
 
\begin{itemize}
    \item We introduce \textbf{MedUnifier}, a novel Med-VLP framework that unifies the current VLP paradigm with a language-guided visual generation task, marking a significant step toward an all-in-one VLP model that seamlessly integrates visual and linguistic information.
    \item We designed a novel TIG module to capture fine-grained details by recovering pixel-level information from hierarchical multi-modal representations, enabling the model to identify subtle visual details, commonly available in medical data (e.g. small nodules, slight opacities, etc).
    \item We perform a series of experiments on various downstream studies using Chest X-rays, showcasing performance enhancements over existing methods across uni-modality, cross-modality, and multi-modality tasks.
    \item We also demonstrate the model's adaptability in generating realistic medical images and reports, highlighting its unique capability to augment out-of-distribution datasets.
    
\end{itemize}

%% file: sec/2_related.tex
\section{Related Work}
\label{sec:related}

\subsection{Vision-and-language Pre-training (VLP)}
Vision-language models (VLMs) have attracted considerable attention due to their powerful ability to integrate visual and textual data, significantly enhancing image captioning, visual question answering, and cross-modal retrieval. The predominant paradigms for VLP can be broadly classified into two main categories. The first one focuses on learning uni-modal encoders for text and images~\cite{jia2021scaling, zhang2022contrastive, huang2021gloria, zhou2022generalized, you2023cxr, wu2023medklip, wang2022medclip}, respectively. However, this dual-encoder architecture limits the capacity to establish intricate interactions between text and image. Another line of work predominantly focuses on fusion encoder-based structure~\cite{wang2022image, bao2022vlmo, su2019vl, li2021align} to facilitate meaningful interactions between the two modalities. In the medical setting, Chen~\etal~\cite{chen2023towards} proposed an effective framework to unify dual-encode style and fusion-encoder. However, these methods do not take the generation of visual information into consideration, and lack exploration of detailed vision content. In this study, we adopt and extend a fusion encoder-based framework to better align visual and textual features by incorporating vector quantization to enable the learning of discrete visual representations, thus facilitating effective vision generation guided by pertinent textual information. Compared to existing studies, our work aligns various modalities and simultaneously creates generic and versatile representations by leveraging the complementary strengths of various losses synergistically, therefore alleviating the additional pre-training stages for expert image tokenizer and iterative denoising.




\begin{figure*}[tp]
    \centering{\includegraphics[width=\linewidth]{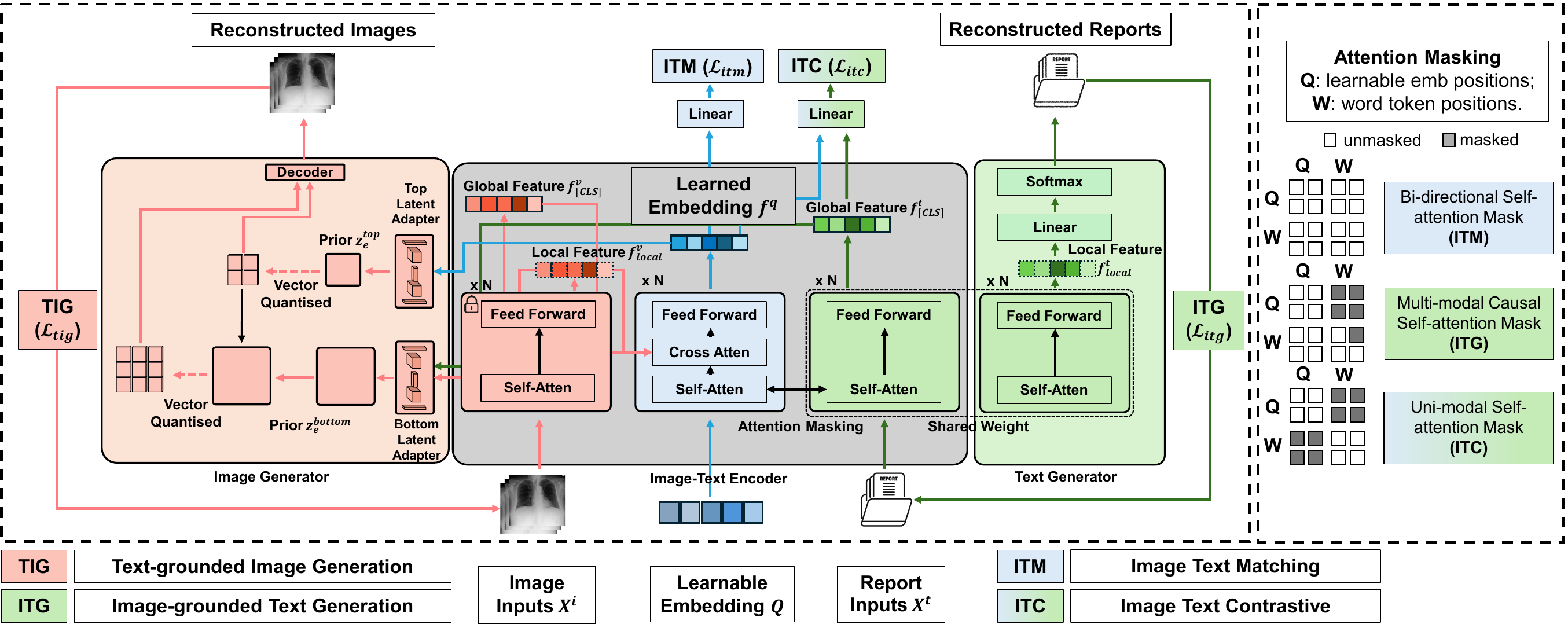}}
    \caption{\textbf{Left}: model architecture consists of an image-text encoder, a text generator, and an image generator to extract the most relevant visual and textual representations by optimizing four distinctive loss functions (ITM, ITC, ITG, TIG). \textbf{Right}: self-attention masking strategies for different learning objectives. \textbf{Bottom}: detailed learning objectives. Integrating visual and textual information enables deep fusion through cross-modal interaction and allows each modality to be processed independently for uni-modal generation. }
    \label{fig2}
\end{figure*}

\subsection{Text to Image (T2I) Generation}
Text-to-image (T2I) tasks aim to generate an image according to a given textual description. Generative adversarial networks (GANs)~\cite{kang2023scaling, li2022stylet2i, reed2016generative, zhang2021cross} and auto-regressive (AR) transformers~\cite{ding2022cogview2, gafni2022make, ramesh2021zero, yu2022scaling} were widely recognized for their exceptional performance and popularity. Advancements in variational auto-encoders (VAEs) and vector quantized VAEs (VQ-VAEs) have further improved T2I generation by introducing a discrete latent space for more stable generation~\cite{gu2022vector, ramesh2021zero, yu2022scaling, ding2021cogview, nichol2021glide}. LLM-CXR \cite{lee2023llm} incorporated pre-trained VQ-GAN into powerful LLM, predicting dual-modal tokens. Moreover, diffusion models~\cite{dhariwal2021diffusion, ho2020denoising} have recently taken the leading position in T2I generation tasks. By adapting the advanced diffusion model, MedM2G \cite{zhan2024medm2g} emphasized high-quality content generation. Despite their effectiveness, diffusion models are notably resource-intensive, often necessitating thousands of iterative steps for denoising, leading to significantly slower speeds. It is hard for diffusion models to get evident visual features due to modelling data distribution implicitly \cite{song2020denoising}. Consequently, for our T2I tasks, we turn to VQ-VAEs to learn more robust representations, thereby enhancing the quality and efficiency of medical image generation effectively.

%% file: sec/3_method.tex
\section{Method}
\label{sec:method}

In this section, we introduce our \textbf{MedUnifier} framework for aggregating four key learning objects on Med-VLP. We first formulate the problem to be solved in \S\ref{subsec:problem} . Then we describe the process of extraction and fusion of multi-modal features using the base model in \S\ref{subsec:enc} with model architecture presented in \S\ref{subsec:model}. Lastly, we illustrate the integration of our proposed text-grounded image generation module and connection with the base model in \S\ref{subsec:bridge}.

\subsection{Problem definition}
\label{subsec:problem}
We formulate the Med-VLP problem with inspiration from the previous studies \cite{chen2023towards, chen2020generating}. Formally, given a set of medical images $X^{I} \in \left\{x^i_{1}, x^i_{2}, \dots ,x^i_{n}\right\}$ with corresponding clinical reports $X^{T} \in \left\{x^t_{1}, x^t_{2}, \dots ,x^t_{n}\right\}$, the entire pre-training objective function can be defined as 
\begin{align}
      \mathcal{L}_{total}  &= \sum_{m=1}^{M} \lambda_{m} \mathcal{L}_{m} (\mathcal{H}_{m}(\mathcal{F}(X^{I}, X^{T})))
\end{align}
where $\mathcal{F}$ represents backbone taking the paired $[x^i, x^t]$ as input. $\mathcal{H}_m$ stands for task-specific modules for further encoding visual and textual features. $\mathcal{L}_m$ and $\lambda_m$ are different loss functions and their weights for the overall loss calculation with the total number of loss functions being $M$. 

\subsection{Model architecture}
\label{subsec:model}
The proposed model mainly consists of three components, an image-text encoder, a text generator, and an image generator with cross-attention layer, masking strategies and vector discretization for extraction on the visual and textual representations, as shown in Figure \ref{fig2}.


\paragraph{Image-text encoder:} We adopted BERT-styled Transformer \cite{pmlr-v202-li23q} as image-text encoder network for fusing multi-modal information in depth. The image-text encoder is constructed by 12 transformer blocks. Its input contains a set of learnable embeddings and clinical reports tokenized by words \cite{devlin2018bert}. Learnable embeddings and textual input interact with each other through self-attention layers \cite{vaswani2017attention} using different masking strategies for various pre-training tasks. We further processed input images into a set of patch embeddings by using a pre-trained, frozen Vision Transformer (ViT) from \cite{fang2023eva}, attaining informative visual features while saving computational costs during training. Moreover, the initial visual embeddings engage with the image-text encoder network through cross-attention layers, which are integrated within each transformer block. At the final layer of the last transformer block, we incorporated distinct, task-specific heads tailored to address various tasks. 

\paragraph{Text generator:} To generate accurate and coherent medical reports, we duplicated the text encoder of the image-text encoder as a language-generative network with shared weights (coloured in green). A decoding head was further added on the top to map each word token embedding to the vocabulary dictionary.

\paragraph{Image generator:} The primary objective of VLP is to learn representations that are transferable across tasks. While the most advanced vision generative models, such as GANs and DMs, excel at generating high-quality and diverse visual content, they typically capture data distributions implicitly, which these models make it challenging to explicitly access intermediate visual representations, limiting their utility in VLP applications. Furthermore medical data often display similar visual or textual patterns and hence encompass nuanced yet crucial medical pattern, which is often overlooked by the existing VLM approaches. To address these constraints, we integrated a vector-quantized variational auto-encoder (VQ-VAE) \cite{van2017neural} within a cross-modal interactive fusion framework to generate high-quality synthetic visual content and improve performance across multiple downstream tasks. 


\subsection{Fusion of visual and textual features}
\label{subsec:enc}
Here, we outline the information flow within the image-text encoder, as well as describe three types of pre-training tasks (ITC, ITM, ITG) in detail.

Given that $x^i \in \mathbb{R}^{C \times H \times W}$,  we divided the entire image into $L_v$ patches with spatial size $(h, w)$ through convolutional operation and added learnable positional encodings:
\begin{align}
    \boldsymbol{X}^i = [\boldsymbol{p}_{[CLS]}, \boldsymbol{p}_1, \boldsymbol{p}_2, \dots, \boldsymbol{p}_{L_v}] + \boldsymbol{E}^{v}_{pos}
\end{align}
We prepended $\boldsymbol{p}_{[CLS]}$ for aggregation of visual information. Then these patch embeddings get passed through a standard pre-trained ViT-g, denoted as $E_I$, to attain preliminary visual embeddings $\boldsymbol{f}^{v}\in \mathbb{R}^{(L_{v}+1)\times d_{v}}$:
\begin{align}
    E_{I}(\boldsymbol{X}^i) = \boldsymbol{f}^{v} &= [\boldsymbol{f}^{v}_{[CLS]}, \boldsymbol{f}^{v}_{local}]\\
    &= [\boldsymbol{f}^{v}_{[CLS]}, \boldsymbol{f}^{v}_1, \boldsymbol{f}^{v}_2, \dots, \boldsymbol{f}^{v}_{L_v}]
\end{align}
where $\boldsymbol{f}^{v}_{[CLS]}$ is global visual feature,  $\boldsymbol{f}^{v}_{local} \in \mathbb{R}^{L_v \times d_v}$ represent local visual features.

For the textual input, we followed BERT \cite{devlin2018bert} to tokenize the input text to word embeddings, adding learnable positional encodings:
\begin{align}
    \boldsymbol{X}^t = [\boldsymbol{w}_{[SPE]}, \boldsymbol{w}_1, \boldsymbol{w}_2, \dots, \boldsymbol{w}_{L_t}] + \boldsymbol{E}^{t}_{pos}
\end{align}
We prepended a special token $\boldsymbol{w}_{[SPE]}$, utilizing different $[SPE]$ tokens for identifying differing tasks.

In order to enable the interaction between word embeddings and preliminary visual embeddings, we constructed a set of learnable embeddings, denoted as $\boldsymbol{Q} = [\boldsymbol{q}_1, \boldsymbol{q}_2, \dots, \boldsymbol{q}_{L_q}], \boldsymbol{Q}\in \mathbb{R}^{L_q\times d_{q}}$. We unified word embeddings and learnable embeddings of the same feature dimension, e.g. $d_t = d_q$. Then we concatenated $\boldsymbol{Q}$ and $\boldsymbol{X}^t$ to form the input of the image-text encoder, denoted as $E_Q$, encoding it to get output embeddings:
\begin{align}
    E_{Q}([\boldsymbol{Q}, \boldsymbol{X}^t]) &= [\boldsymbol{f}^{q}, \boldsymbol{f}^{t}] \\
    &= [\boldsymbol{f}^{q}, \boldsymbol{f}^{t}_{[SPE]}, \boldsymbol{f}^{t}_{local}]
\end{align}
We employed a cross-attention mechanism \cite{alayrac2022flamingo} to facilitate interaction between learnable embeddings and preliminary visual embeddings. This design enables $\boldsymbol{f}^q$ to function as the final visual representation. To ensure clarity over the fusion of learnable embeddings and textual representations, we implemented distinct masking strategies within the self-attention layers (Figure \ref{fig2}, right panel). 

\paragraph{Image-text contrastive learning (ITC)} The task aims to align visual and textual representations by maximizing their mutual information through a contrastive approach. To accomplish this, we replaced $\boldsymbol{w}{[SPE]}$ with $\boldsymbol{w}{[CLS]}$ to facilitate global textual representations denoted as $\boldsymbol{f}_{[CLS]}^t \in \mathbb{R}^{d_t}$. Furthermore, we implemented uni-modal masking (Figure \ref{fig2}, right panel) to enable learnable embeddings $\boldsymbol{Q}$ and textual embeddings $\boldsymbol{X}^{t}$ to attend exclusively to themselves.$\boldsymbol{f}^{q}$ and $\boldsymbol{f}_{[CLS]}^t$ are then linearly projected to representations as:


\begin{align}
    \boldsymbol{g}^{q} &= \mathcal{H}_{itc}^{q}(\boldsymbol{f}^{q})\\
    \boldsymbol{g}^{t} &= \mathcal{H}_{itc}^{t}(\boldsymbol{f}^{t}_{[CLS]})
\end{align}
where $\mathcal{H}_{itc}^{q}, \mathcal{H}_{itc}^{t}$ are ITC heads. We computed the pairwise similarity between each visual and textual representation $\boldsymbol{g}^{q}$ and $\boldsymbol{g}^{t}$ and chose the highest one as the image-text similarity to calculate bi-directional contrastive loss:
\begin{align}
    \mathcal{L}^{(q|t)}_{itc} = \frac{1}{N}\sum_{k=1}^{N}-\log(\frac{\exp(\max<\boldsymbol{g}^{q}_{k}, \boldsymbol{g}^{t}_{k}> / \tau)}{\sum_{n=1}^{N}\exp(\max<\boldsymbol{g}^{q}_{k}, \boldsymbol{g}^{t}_{n}> / \tau)})\\
    \mathcal{L}^{(t|q)}_{itc} = \frac{1}{N}\sum_{k=1}^{N}-\log(\frac{\exp(\max<\boldsymbol{g}^{q}_{k}, \boldsymbol{g}^{t}_{k}> / \tau)}{\sum_{n=1}^{N}\exp(\max<\boldsymbol{g}^{q}_{n}, \boldsymbol{g}^{t}_{k}> / \tau)})
\end{align}
where $\tau \in \mathbb{R}$ is a scaling temperature parameter initialized to 0.07, $N$ is mini-batch size and $⟨\cdot, \cdot⟩$ represents the cosine similarity. The overall ITC loss is defined as:
\begin{align}
    \mathcal{L}_{itc} =  \frac{1}{2}(\mathcal{L}^{(q|t)}_{itc} + \mathcal{L}^{(t|q)}_{itc})
\end{align}
We here expanded visual representation space from a conventional single vector to a set of vectors e.g. $\boldsymbol{f}^{q}\in \mathbb{R}^{L_q\times d_{q}}$  which is different from \cite{huang2021gloria, wu2023medklip}.

\paragraph{Image-text matching (ITM)} This task aims to learn a precise alignment by classifying image-text pairs as either positive or negative. We implemented a bi-directional mask (Figure \ref{fig2}, right panel) that allows all learnable embeddings and word token embeddings to attend to one another. The resulting output of learned embeddings denoted as $\boldsymbol{f}^{q}$, capture enriched multi-modal information. These tokens were then fed into a two-class linear classifier, $\mathcal{H}_{itm}$, where the outputs are averaged across learned embeddings to generate a logit and compute the Image-Text Matching (ITM) loss:
\begin{align}
    \mathcal{L}_{itm} &= \frac{1}{N}\sum_{k=1}^{N} -\log(p(Y_k|\hat{Y_k})) \\
    \hat{Y} &= \frac{1}{L_q}\sum_{i=1}^{L_q}\mathcal{H}_{itm}(\boldsymbol{f}^{q}_i),
\end{align}
$Y$ represents ground truth labels within mini-batch by hard negative samples mining, as stated in \cite{li2021align}.

\paragraph{Image-grounded text generation (ITG)} This task is to generate text conditioned on paired images. To achieve a coherent and precise generation of medical reports within a unified VLP framework, we chose CLM \cite{radford2018improving, brown2020language} where each word token attends only to preceding tokens, following a GPT-style language model architecture \cite{brown2020language}. Inspired by UniLM \cite{dong2019unified}, we implemented a multi-modal causal self-attention mask (Figure \ref{fig2}, right panel). We replaced the special token $\boldsymbol{w}{[SPE]}$ with $\boldsymbol{w}{[DEC]}$ to signal a decoding task. We also introduced a word prediction head, denoted as $\mathcal{H}_{itg}$. This learning objective is formalized as:
\begin{align}
    \mathcal{L}_{itg} &= \frac{1}{N L_t}\sum_{k=1}^{N} \sum_{i=1}^{L_t}-\log (p_i) \\
    p_i &= Softmax(\mathcal{H}_{itg}(\boldsymbol{f}^t_{local}))\\
    &= p(\boldsymbol{w}_{i}|\boldsymbol{Q}, \dots, \boldsymbol{w}_{i-1})
\end{align}

\subsection{Text-grounded image generation (TIG)}
\label{subsec:bridge}

We also introduce an innovative and efficient module designed for the text-grounded image generation task, with integration into the previously discussed image-text encoder and text generator for a versatile Med-VLP model. 

\paragraph{Rethinking VQ-VAE} The VQ-VAE \cite{van2017neural} offers a significant advantage over other generative models due to its ability to explicitly learn discrete visual representations. This feature aligns closely with the image-text encoder, which similarly learns discrete representations from a dictionary of learnable embeddings. Motivated by this, we chose to adopt VQ-VAE as the image generator in our framework, which forms a different unified Med-VLP framework from other studies \cite{zhou2022advancing, chen2023contrastive, chen2023towards}. Additionally, the image-text encoder generates two distinct types of visual features: one for abstract visual representations, denoted as $\boldsymbol{f}^q$, and another for fine-grained, local visual embeddings, $\boldsymbol{f}^v_{local}$. Consequently, the image-text encoder can be viewed as a powerful multi-modal encoder, akin to the image encoder in conventional VAEs \cite{van2017neural, higgins2017beta}. Inspired by the work of \cite{razavi2019generating}, we developed hierarchical vector quantizers and image decoders. Figure \ref{fig2} further illustrates the proposed text-grounded image generator. Our TIG module is specifically designed to capture fine-grained details through recovering pixel-level information from hierarchical multi-modal representations, enabling the model to identify subtle visual details. 

\paragraph{Bridging the gap}
The features $\boldsymbol{f}^q \in \mathbb{R}^{L_q \times d_q}$ contains textual implications derived from the co-training image-text encoder, which we regard as the top latent representation, denoted $\boldsymbol{z}^{top} \in \mathbb{R}^{L_q \times d_q}$. In contrast, $\boldsymbol{f}^v_{local} \in \mathbb{R}^{L_v \times d_v}$ does not include textual information. Therefore, we concatenate $\boldsymbol{f}^v_{local}$ with the aggregated textual representation, $\boldsymbol{f}^{t}_{[CLS]} \in \mathbb{R}^{1 \times d_t}$, along the feature dimension, resulting in the multi-modal bottom latent representation $\boldsymbol{z}^{bot} \in \mathbb{R}^{L_v \times (d_v + d_t)}$. 

At the top level, we devise the latent adapter, denoted as $\mathcal{Z}_{top}$, for transforming $\boldsymbol{z}^{top}$ into spatial feature map $\boldsymbol{z}_{e}^{top}$:
\begin{align}
    \boldsymbol{z}_{e}^{top} = \mathcal{Z}_{top}(\boldsymbol{z}^{top})
\end{align}
where $\mathcal{Z}_{top}$ consisted of a nonlinear transformation, spatial positional encoding summer \cite{wang2019translating} and a residual block \cite{he2016deep}. The illustration can be found in appendix \ref{sec:arc_la}. We reshaped $\boldsymbol{z}^{top}$ and then pass it through $\mathcal{Z}_{top}$, reaping $\boldsymbol{z}_{e}^{top} \in \mathbb{R}^{d_{top} \times h_{top} \times w_{top}}$. Followed by a vector quantization layer with a latent embedding space $e^{top}$, we gain discrete feature map $\boldsymbol{z}_{q}^{top} \in \mathbb{R}^{d_{top} \times h_{top} \times w_{top}}$:
\begin{align}
    \boldsymbol{z}_{q}^{top} = quantizer_{top}(\boldsymbol{z}_{e}^{top})
\end{align}

At the bottom level, in accordance with the top level, a latent adapter and vector quantizer with a latent embedding space $e^{bottom}$ are deployed to gain a discrete feature map:
\begin{align}
    \boldsymbol{z}_{e}^{bot} &= \mathcal{Z}_{bot}(\boldsymbol{z}^{bot})\\
    \boldsymbol{z}_{q}^{bot} &= quantizer_{bot}(\boldsymbol{z}_{e}^{bot}, \boldsymbol{z}_{q}^{top})
\end{align}
where $\boldsymbol{z}^{bot}$ are viewed as having spatial size $h_{bot} \times w_{bot} = \frac{H}{h} \times \frac{W}{w}$. Producing $\boldsymbol{z}_{q}^{bottom}$ is conditioning on $\boldsymbol{z}_{q}^{top}$. 

We built hierarchical decoders $\mathcal{D}$ to recover raw images from discrete multi-modal representations:
\begin{align}
    \hat{x}^i = \mathcal{D}(\boldsymbol{z}_{q}^{top}, \boldsymbol{z}_{q}^{bot})
\end{align}
The text-grounded image generation (TIG) loss is formulated as:
\begin{align}
    \mathcal{L}_{tig} &= \frac{1}{N}\sum_{k=1}^{N}-\log p(x^i_k|\boldsymbol{z}_{q}^{top}, \boldsymbol{z}_{q}^{bot}) \\ \notag
    &+\left\Vert \text{sg}[\boldsymbol{z}_{e}^{top}]-e^{top} \right\Vert_2 +\beta_1 \left\Vert \text{sg}[e^{top}]-\boldsymbol{z}_{e}^{top} \right\Vert_2\\ \notag
    &+\left\Vert \text{sg}[\boldsymbol{z}_{e}^{bot}]-e^{bot} \right\Vert_2 +\beta_2 \left\Vert \text{sg}[e^{bot}]-\boldsymbol{z}_{e}^{bot} \right\Vert_2
\end{align}
where the negative logarithmic term can be written as mean square error (MSE) $\left\Vert x^i_k - \hat{x}^i_k\right\Vert_2$, $\text{sg}[\cdot]$ is gradient stop operation. Hyper-parameters $\beta_1, \beta_2$ are both set to be 0.5.

\subsection{Total learning objectives}

We provide a comprehensive summary of all the learning objectives and present the ultimate loss function:
\begin{align}
    \mathcal{L}_{total} = \lambda_1\mathcal{L}_{itc} + \lambda_2\mathcal{L}_{itm} + \lambda_3\mathcal{L}_{itg} + \lambda_4\mathcal{L}_{tig}
\end{align}
Four weights $\lambda$ were set to 1 in experiments. These weights were determined by ablation study (see appendix \ref{sec:more abl}).

\begin{table*}[t!]
\centering
\begin{tabular}{@{}lccccccccc@{}}
\toprule
\multirow{3}{*}{Methods} & \multicolumn{3}{c}{RSNA (AUC)} & \multicolumn{3}{c}{SIIM (AUC)} & \multicolumn{3}{c}{COVIDx (ACC)} \\ \cmidrule(l){2-10} 
\multicolumn{1}{c}{}                                 & 1\%      & 10\%     & 100\%    & 1\%      & 10\%     & 100\%    & 1\%       & 10\%     & 100\%     \\ \midrule
ConVIRT\cite{zhang2022contrastive}                                              & 84.2     & 86.9     & 88.7     & 84.2     & 85.7     & 91.5     & 72.5      & 82.5     & 92.0      \\
GLoRIA\cite{huang2021gloria}                                               & 84.1     & 86.8     & 89.1     & 85.1     & 88.5     & 92.1     & 66.5      & 80.5     & 88.0      \\
BioViL\cite{boecking2022making}                                               & 82.0     & 85.4     & 88.6     & 79.8     & 81.6     & 90.5     & -         & -        & -         \\
LoVT \cite{muller2022joint}                                                & 85.1     & 86.5     & \underline{89.3}     & 85.5     & 88.5     & 92.2     & -         & -        & -         \\
MGCA \cite{wang2022multi}                                                & 85.8     & \underline{87.7}     & 89.2     & 86.1     & \underline{89.6}     & 92.0     & 74.8      & 84.8     & \underline{92.3}      \\
PRIOR \cite{cheng2023prior}                                              & 85.7     & 87.1     & 89.2     & 87.2     & 89.1     & 92.3     & -         & -        & -         \\ \midrule
MedUnifer (w/o TIG)                                 & \underline{86.3}     & 87.1     & 88.1     & \underline{87.6}     & 88.7     & \underline{92.3}     & \underline{75.3}      & \underline{87.5}     & \underline{92.8}      \\
MedUnifer (w TIG)                                   & \textbf{87.6}     & \textbf{88.8}     & \textbf{91.7}    & \textbf{87.9}     & \textbf{92.5}     & \textbf{94.8}     & \textbf{76.8}      & \textbf{88.3}    & \textbf{93.5}      \\ \bottomrule
\end{tabular}
\caption{
Fine-tuned image classification results on RSNA, SIIM and COVIDx with 1\%, 10\%, 100\% training data. Area under ROC curve (AUROC [\%]) are reported for RSNA and SIIM datasets, and accuracy (ACC [\%]) is reported for COVIDx dataset. The best and second-best results are highlighted in bold and underlined, respectively. Our method achieves the best performance across all datasets.}
\vspace{-0.05in}
\label{tab:fine-tuning}
\end{table*}

\begin{table}[t!]
\centering
\resizebox{\columnwidth}{!}{
\begin{tabular}{@{}lcccccc@{}}
\toprule
\multirow{2}{*}{Methods} & \multicolumn{3}{c}{Image-Text Retrieval (ITR)} & \multicolumn{3}{c}{Text-Image Retrieval (TIR)} \\ \cmidrule(l){2-7} 
                         & mAP@1  & mAP@5 & mAP@10 & mAP@1  & mAP@5 & mAP@10 \\ \midrule
ConVIRT\cite{zhang2022contrastive}                  & 46.5   & 53.9  & 53.8   & 20.0   & 45.4  & 35.5   \\
GLoRIA\cite{huang2021gloria}                   & 46.7   & 56.4  & 55.0   & 51.8   & 59.5  & 58.9   \\
BioViL\cite{boecking2022making}                   & 47.3   & 57.7  & 55.6   & 54.6   & 64.3  & 62.8   \\
MedCLIP \cite{wang2022medclip}                 & 47.6   & 58.0  & 55.9   & 56.3   & 69.9  & \underline{66.7}   \\
MGCA \cite{wang2022multi}                    & 47.1   & 57.4  & 55.4   & 53.1   & 61.9  & 61.1   \\
REFERS \cite{zhou2022generalized}                  & 52.4   & 59.9  & 58.6   & \underline{60.6}   & \textbf{71.9}  & \textbf{69.0}   \\
CXR-CLIP \cite{you2023cxr}                & 51.8   & 61.2  & 58.5   & 60.2   & 69.2  & 64.6   \\ \midrule
MedUnifier (w/o TIG)     & \underline{57.4}   & \underline{65.4}  & \underline{60.3}   & 59.6   & 68.3  & 63.4   \\
MedUnifier (w TIG)       & \textbf{60.7}   & \textbf{66.6}  & \textbf{61.7}  & \textbf{63.1}   & \underline{70.8}  & 64.4   \\ \bottomrule
\end{tabular}
}
\caption{
Cross-modal retrieval results on MIMIC-CXR 5x200 dataset. The top K (1, 5, 10) mean Average Precision metrics are reported. Our method achieves the best performance for ITR tasks.
}
\vspace{-0.05in}
\label{tab:Retrival}
\end{table}

\begin{table}[t!]
\centering
\resizebox{\columnwidth}{!}{
\begin{tabular}{@{}l|c|c|c@{}}
\toprule
Methods &
  \begin{tabular}[c]{@{}c@{}}MIMIC 5x200 \\ACC\end{tabular} &
  \begin{tabular}[c]{@{}c@{}}CheXpert 5x200 \\ACC\end{tabular} &
  \begin{tabular}[c]{@{}c@{}}RSNA\\ ACC\end{tabular} \\ \midrule
ConVIRT\cite{zhang2022contrastive}              & 43.8 & 35.2 & 77.4 \\
GLoRIA\cite{huang2021gloria}                & 47.5 & \textbf{45.0} & 68.3 \\
BioViL\cite{boecking2022making}               & 48.5 & 42.2 & 77.1 \\
MedCLIP\cite{wang2022medclip}               & 47.1 & 41.1 & 81.8 \\
MGCA \cite{wang2022multi}                & 48.0 & 40.9 & 76.2 \\
REFERS \cite{zhou2022generalized}               & 49.5 & 41.8 & 78.0 \\
CXR-CLIP \cite{you2023cxr}             & \underline{49.7} & 35.9 & 76.9 \\ \midrule
MedUnifer (w/o TIG) & 44.8 & 40.8 & \textbf{85.0} \\
MedUnifer (w TIG)   & \textbf{50.4 }& \underline{43.5} & \underline{82.0} \\ \bottomrule
\end{tabular}
}
\caption{
Zero-shot image classification results on MIMIC 5x200, CheXpert 5x200 and RSNA datasets. Our method achieves the best performance for MIMIC 5x200 and RSNA datasets. Note that GLoRIA is trained on the CheXpert dataset.}
\vspace{-0.05in}
\label{tab:ZS}
\end{table}

%% file: sec/4_experiments.tex
\section{Experiments}
\label{sec:experiments}
We perform the pre-training on the current largest multi-modal medical dataset, MIMIC-CXR v2.0.0 \cite{johnson2019mimic} and evaluate our model on various downstream tasks, followed by an ablation study for probing purpose, which shows the proposed method's superiority.

\subsection{Implementation details}

We employed a BERT model as the primary network for the image-text encoder and utilized ViT-g as the pre-trained ViT. The input image resolution was set to $224 \times 224$, with a maximum text length of 95 tokens, and 32 learnable embeddings. The top and bottom codebook size were both set to 512 with feature dimension of 768. For optimization, we applied the AdamW optimizer \cite{loshchilov2017decoupled} with parameters $\beta_1 = 0.9$, $\beta_2 = 0.95$, and a weight decay of 0.05. A cosine learning rate decay schedule was used, with a peak learning rate of 1e-4. We incorporated a warm-up phase for the initial 5\% of training steps, starting with a learning rate of 1e-5. The pre-training process was conducted on four NVIDIA A100 GPUs.The implementation details are further elaborated in appendix \ref{sec:imp}.


\subsection{Medical Vision-and-Language Benchmark}
To assess the effectiveness of our method, we conducted experiments across three types of tasks: uni-modal, cross-modal, and multi-modal. We conducted experiments on MedUnifier with/without TIG and compared with previous studies as our main experimental results. All benchmark datasets used in experiments pertain specifically to radiology with detailed descriptions in the appendix \ref{sec:data}. For baselines, we took reference from papers and implemented them using official codes: ConVIRT\cite{zhang2022contrastive}, GLoRIA\cite{huang2021gloria}, BioViL\cite{boecking2022making}, LoVT \cite{muller2022joint}, MedCLIP\cite{wang2022medclip}, MGCA \cite{wang2022multi}, PRIOR \cite{cheng2023prior}, REFERS \cite{zhou2022generalized}, CXR-CLIP \cite{you2023cxr}, ViLT\cite{bannur2023learning}, R2Gen \cite{chen2020generating}, PTUnifier \cite{chen2023towards}, DCL \cite{li2023dynamic}, MOTOR \cite{lin2023towards}, UniXGer\cite{lee2023vision}, RoentGen\cite{chambon2022roentgen}, LLM-CXR\cite{lee2023llm}.


\vspace{12pt}

\noindent \textbf{Uni-modal tasks} ~ ~ assess the learned visual representations for image modality by applying uni-modal masking within classification scope on datasets such as RSNA Pneumonia \cite{shih2019augmenting}, SIIM-ACR \cite{siim-acr-pneumothorax-segmentation}, and COVIDx \cite{Wang2020}. To examine the model's data efficiency, we fine-tune it using different proportions of the training data (1\%, 10\%, or 100\%).

\paragraph{Cross-modal tasks} require models to align vision and language modalities. We conduct experiments across three tasks: image-to-text retrieval (ITR), text-to-image retrieval (TIR), and zero-shot image classification (ZS). For ITR and TIR, we report cross-modal information retrieval metrics, the mean Average Precision at K (mAP@K), using the MIMIC 5x200 dataset. In the ZS task, we employ MIMIC 5x200, CheXpert 5x200 \cite{huang2021gloria, cheng2023prior}, and draw 500 positive and 500 negative samples from the full RSNA Pneumonia dataset \cite{shih2019augmenting} for evaluation purpose.

\paragraph{Multi-modal tasks} generate uni-modal content through multi-modal interaction. We carry out two kinds of experiments including image-grounded text generation (ITG) and text-grounded image generation (TIG). For ITG, we adopt the MIMIC-CXR held-out test set to evaluate the quality of generated reports. Standard natural language generation (NLG) criteria are used to assess the performance, including BLEUn \cite{papineni2002bleu}, METEOR \cite{denkowski2011meteor}, and ROUGE-L \cite{lin2004rouge}. For TIG, following \cite{razavi2019generating}, we first train two Pixelsnail models \cite{chen2018pixelsnail} to model multi-modal priors $\boldsymbol{z}_{q}^{top}, \boldsymbol{z}_{q}^{bot}$. Then we sample latent encodings from both priors and generate new medical images by decoding latent encodings. For quantitative analysis, we present FID \cite{heusel2017gans} scores.

\subsection{Results and Analyses}
\begin{figure}[h]
    \centering{\includegraphics[width=\linewidth]{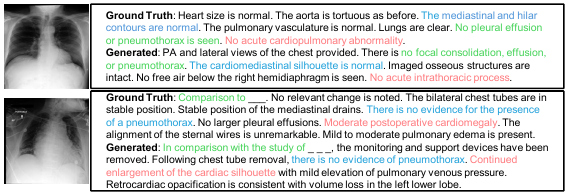}}
    \caption{Comparison of ground truth and generated radiology reports reveals strong semantic alignment. In the top figure, both reports describe normal heart size, no pneumothorax or pleural effusion, and a normal cardiomediastinal silhouette, with the generated text adding details on osseous structures/intrathoracic processes. In the bottom figure, both reports align on pneumothorax and cardiomegaly. The same colours denote matched content between the generated sequences and the ground truth report.}
    \label{fig4}
\end{figure}

To validate the effectiveness of MedUnifier, we conduct experiments on the above vision-and-language benchmark. The results of the main experiments are presented in Table \ref{tab:fine-tuning},\ref{tab:Retrival}, \ref{tab:ZS}, \ref{tab:NLG} \ref{tab:Fid} and Figure \ref{fig3}. We observe several noteworthy findings in our results. First, our model outperforms prior studies on uni-modal tasks across various downstream datasets, as shown in Table \ref{tab:fine-tuning}. This improvement suggests that integrating TIG significantly enhances the model's ability to learn more transferable visual representations. Second, for cross-modal retrieval, MedUnifier model achieves the highest performance, demonstrating a superior ability to understand and integrate cross-modal content compared to other models (see Table \ref{tab:Retrival}). The obtained results further highlight the better performance of MedUnifier, suggesting that our approach supports the necessary complementary semantic data for cross-modal retrieval. We also observe that excluding the TIG module slightly decreases MedUnifier's effectiveness in text-image retrieval tasks, which may be a result of an over-reliance on ITG that leads to an imbalance in vision-language fusion. In addition, MedUnifier demonstrates better performance on zero-shot classification tasks for both the MIMIC 5x200 and RSNA datasets in Table \ref{tab:ZS}. However, GLoRIA outperforms our model on the CheXpert 5x200 dataset, likely due to its pre-training on the full CheXpert dataset with accompanying medical reports.  Nonetheless, the results indicate the effectiveness of employing prompt ensembles within the proposed method, leading to enhanced overall performance improvements. Table \ref{tab:NLG} demonstrates that our models, both with and without the TIG module, surpass previous methods for image-grounded medical report generation. The Med-VLP framework also gains substantial advantages from incorporating causal language modelling, which together contributes to its improved performance. In Figure \ref{fig4}, we provide a comparison of the generated report and ground truth report. Finally, we conduct both quantitative and qualitative analyses on text-grounded image generation tasks in Table \ref{tab:Fid} and Figure \ref{fig3} (see appendix). It highlights that MedUnifier with TIG achieve comparable performance for medical vision generation tasks with less model's complexity (one-stage pre-training) then existing models \cite{lee2023llm, lee2023vision, chambon2022roentgen}, all of which require additional pre-training for image tokenizer or iterative denoising. From a direct visual inspection, the reconstructed visual samples are nearly indistinguishable from authentic radiographs. Moreover, synthetic samples generated from multi-modal priors demonstrate high diversity, highlighting their potential to augment out-of-distribution medical data effectively.


\begin{table}[t!]
\centering
\resizebox{\columnwidth}{!}{
\begin{tabular}{@{}l|cccccc@{}}
\toprule
\multirow{3}{*}{Methods} & \multicolumn{6}{c}{\multirow{2}{*}{MIMIC-CXR test set}}                                                  \\
                     & \multicolumn{6}{c}{}                    \\
                         & \multicolumn{1}{l}{BL-1} & \multicolumn{1}{l}{BL-2} & \multicolumn{1}{l}{BL-3} & BL-4 & MTR & RG-L \\ \midrule
                         
DCL\cite{li2023dynamic}       & -    & - & -    & 10.9  & 15.0    & 28.4 \\
MOTOR\cite{lin2023towards}             & -    & 21.3 & -    & \textbf{15.6}  & 19.3    & \underline{31.4}\\
ViLT\cite{bannur2023learning}                 & -    & 21.3 & -    & 9.2  & -    & 29.6 \\
R2Gen \cite{chen2020generating}               & 35.3 & 21.8 & 14.5 & 10.3 & 14.2 & 27.7 \\
PTUnifier \cite{chen2023towards}           & -    & -    & -    & 10.7 & -    & -    \\ \midrule
MedUnifier (w/o TIG) & \underline{40.1} & \underline{24.4} & \underline{16.0} & 11.3 & \underline{18.8} & 29.6 \\
MedUnifier (w TIG)   & \textbf{41.5} & \textbf{26.3} & \textbf{18.1} & \underline{12.9} & \textbf{22.7} & \textbf{34.0} \\ \bottomrule
\end{tabular}
}

\caption{
The performance of all baselines and our method on the test set of MIMIC-CXR dataset for Natural Language Generation (NLG) metrics. BL-n denotes BLEU score using up to n-grams; MTR and RG-L denote METEOR and ROUGE-L, respectively. Our method achieves the best performance across all metrics.
}
\vspace{-0.05in}
\label{tab:NLG}
\end{table}

\begin{table}[t!]
\centering
\small
\begin{tabular}{@{}l|c|l|c@{}}
\toprule
Methods  & FID $\downarrow$  & Methods        & FID $\downarrow$  \\ \midrule
UniXGer\cite{lee2023vision} & 78.2 & Validation set & 17.2 \\
RoentGen\cite{chambon2022roentgen} & \underline{42.4}& Reconstruction & 27.2 \\
LLM-CXR\cite{lee2023llm}  & \textbf{22.8} & MedUnifier     & 46.2 \\ \bottomrule
\end{tabular}%
\caption{
FID score \cite{heusel2017gans} on reconstructed and synthetic images, using the MIMIC-CXR pre-training dataset as the reference dataset. The FID score using MIMIC-CXR validation is also provided for a fair comparison. Reconstructed images are generated by passing the training set through our MedUnifier. Synthetic images are produced from sampled and decoded latent encodings using the trained Pixelsnail models and VAE decoder.
}
\vspace{-0.05in}
\label{tab:Fid}
\end{table}

\begin{table}[t]
\centering
\resizebox{\columnwidth}{!}{
\begin{tabular}{@{}c|cccc|c|c|ccc@{}}
\toprule
\multirow{2}{*}{ID} & \multicolumn{4}{c|}{Learning Objectives} & ITR           & Zero-shot cls                                               & \multicolumn{3}{c}{Fine-tuned cls}                                                                                                                                  \\ \cmidrule(l){2-10} 
                    & ITC      & ITM      & ITG      & TIG     & mAP@1         & \begin{tabular}[c]{@{}c@{}}MIMIC 5x200\\ (ACC)\end{tabular} & \begin{tabular}[c]{@{}c@{}}RSNA\\ (AUC)\end{tabular} & \begin{tabular}[c]{@{}c@{}}SIIM\\ (AUC)\end{tabular} & \begin{tabular}[c]{@{}c@{}}COVID\\ (ACC)\end{tabular} \\ \midrule
1                   &     \checkmark     &          &          &         & 53.7          & 41.4                                                        & 87.0                                                 & 89.4                                                 & 90.8                                                  \\
2                   &     \checkmark     & \checkmark         &          &         & 55.2          & 44.3                                                        & 87.1                                                 & 89.8                                                 & 91.5                                                  \\
3                   &    \checkmark      &  \checkmark        &        \checkmark  &         & 57.4          & 44.8                                                        & 88.1                                                 & 92.3                                                 & 92.8                                                  \\
4                   &    \checkmark      &    \checkmark      &          &   \checkmark      & 58.5          & 46.2                                                        & 89.1                                                 & 92.6                                                 & 91.3                                                  \\
5                   &     \checkmark     &    \checkmark      &       \checkmark   &  \checkmark       & \textbf{60.7} & \textbf{50.4}                                               & \textbf{91.7}                                        & \textbf{94.8}                                        & \textbf{93.5}                                         \\ \bottomrule
\end{tabular}%
}
\caption{
Ablation studies on the different modules. The best performance is achieved using all objectives. Details see appendix \ref{sec:more abl}.
}
\vspace{-0.05in}
\label{tab:Abl}
\end{table}

\subsection{Ablation study}
To demonstrate the efficacy of our proposed method, we perform an ablation study (Table \ref{tab:Abl}) across various learning objectives. The results indicate that using only the ITC loss (ID 1) yields the lowest performance. As ITM and ITG objectives are incrementally incorporated, performance gradually improves (IDs 2 and 3), highlighting how refined cross-modal alignment and multi-modal language modelling enhance the model's overall capabilities. Interestingly, the model with TIG (ID 4) surpasses the one with ITG (ID 3). We attribute this phenomenon to the relative difficulty of generating pixel-level image representations guided by text, as compared to generating word-level representations guided by visual input, which leads the model to learn more abstract, well-aligned representations. Ultimately, the integration of all objective types (ID 5) enables the model to achieve optimal performance, underscoring the viability of incorporating vision generation into existing frameworks.

%% file: sec/5_conclusion.tex
\section{Conclusion}
\label{sec:conclusion}

In this paper, we introduce a novel and unified Med-VLP model, MedUnifier, which optimizes four distinct learning objectives simultaneously. By leveraging learnable embeddings and encoding raw images through a pre-trained Vision Transformer (ViT), MedUnifier circumvents the need to learn visual embeddings from scratch. Additionally, a VQ-VAE-based text-grounded image generation task is further incorporated into the Med-VLP framework to enhance its representation learning capacity. It reconstructs pixel-level visual details from both image and report, facilitating fine-grained visual understanding commonly available in medical data (subtle visual details e.g. small nodules, slight opacities, etc.) and efficient use of multi-modal representations through hierarchical latent adapters of dynamically adjusting the abstraction for each mode. Our proposed method effectively complements existing Med-VLP frameworks and achieves state-of-the-art performance. Our work also has significant implications for enhancing the VLP development for radiological applications.





%% file: sec/6_acknowledgement.tex
\section{Acknowledgement}
This project is supported by the Ministry of Education, Singapore, under its Academic Research Fund Tier 1 (RG25/24 and RS16/23), as well as the Lee Kong Chian School of Medicine - Ministry of Education Start-Up Grant. This research is also partially supported by the computational resources and staff contributions of the Quest high-performance computing facility at Northwestern University, jointly funded by the Office of the Provost, the Office for Research, and Northwestern University Information Technology.

%% file: sec/X_suppl.tex
\clearpage
\setcounter{page}{1}
\maketitlesupplementary



\section{Architecture of latent adapter}
\label{sec:arc_la}
\begin{figure*}[!]
    \centering{\includegraphics[width=\linewidth]{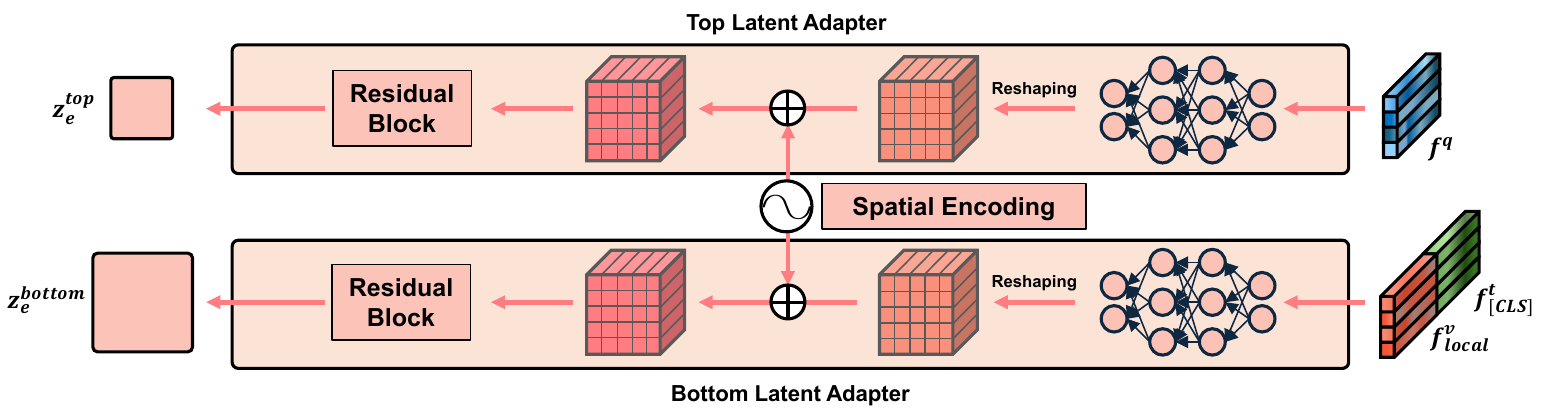}}
    \caption{The detailed model structure of latent adapters. The learned embeddings are fed into top adapter as input while text representation concatenated with local preliminary visual embeddings are put into bottom adapters.}
    \label{figS1}
\end{figure*}

Here, we provide detailed network architecture for latent adapters. We devise a novel latent adapter to transform different levels' multi-modal features into latent representations with its internal structure presented in Figure \ref{figS1}.

\begin{figure*}[tp]
    \centering{\includegraphics[width=\linewidth]{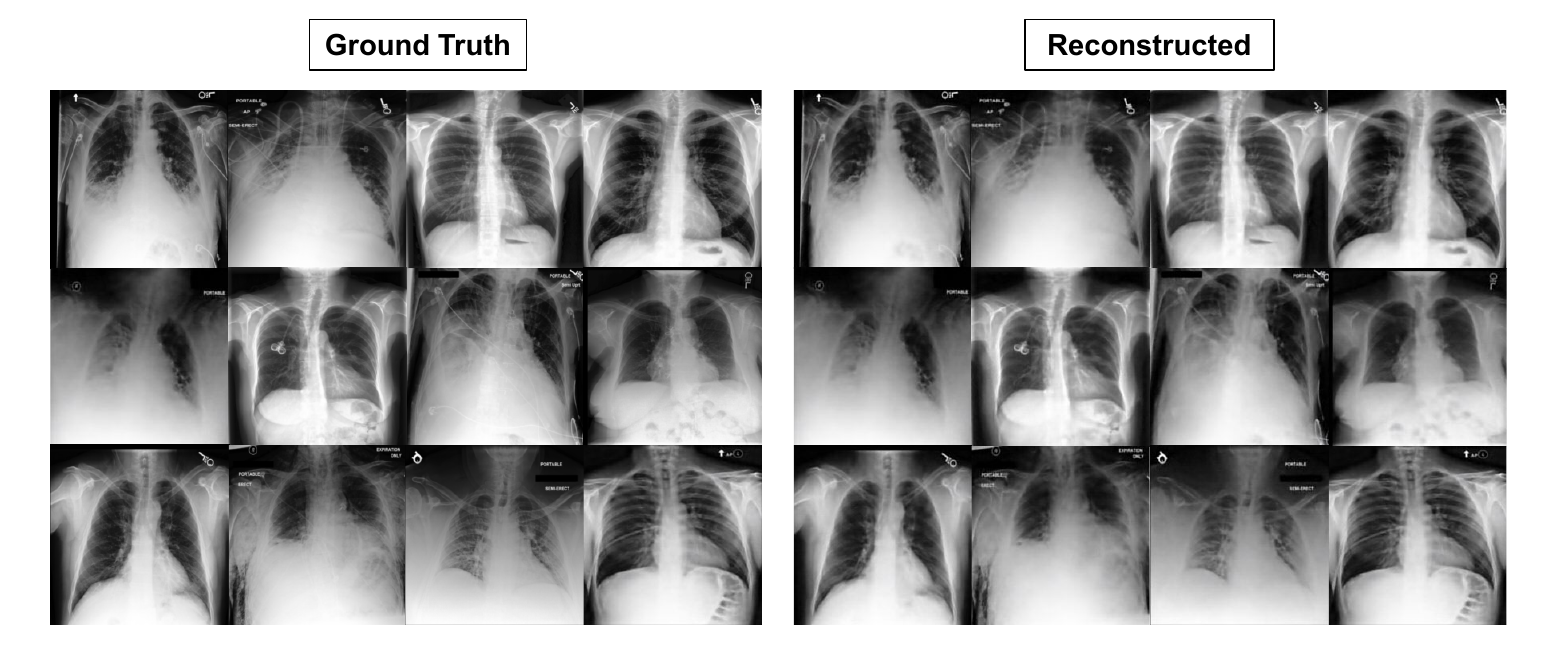}}
    \caption{The additional comparison between real radiographs and reconstructed ones.}
    \label{figS2}
\end{figure*}

\section{Implementation details}
\label{sec:imp}
\paragraph{Pre-training.} For the image-text encoder, we adopt the pre-trained ViT-g from \cite{fang2023eva} as preliminary visual embeddings extractor ($E_I$) and its hidden dimension ($d_v$) is set to 1408. We adopt a base-size Transformer encoder with 12 layers initialized from \cite{boecking2022making} and its hidden dimension ($d_q, d_t$) is set to 768. The length of learnable embeddings was set to 32 ($L_q$). For tokenizing medical reports specialized medicine vocabulary dictionary from \cite{boecking2022making} was employed with vocabulary size of 30522. All text inputs were confined to a maximum length of 95 ($L_t$). All images size was resized into $3 \times 224 \times 224$ ($C\times H\times W$) and normalized into range $(0, 1)$. Raw images were split into non-overlap patches by stride of $14 \times 14$ ($h\times w$) so that patch embeddings length was $16 * 16 = 256$ ($L_v$). The feature dimension of ITC linear heads was set to 256. For the text generator, we set the dimension of the prediction head to 768. For the image encoder, the size of latent feature maps were set to be $8\times 8, 16\times 16$ for top level and bottom level, respectively. For optimization, we adopt the AdamW optimizer \cite{loshchilov2017decoupled} with $\beta_1 = 0.9, \beta_2 = 0.95$, and a weight decay of 0.05. We used a cosine learning rate decay with a peak learning rate of 1e-4. We used the warm-up strategy during the first 5\% of the total number of steps and an initial learning rate of 1e-5. The pre-training process was running on four 80G NVIDIA A100 GPUs. We used the mix precision wuth Accelerate open-source library \cite{accelerate} to speed up training and save computational costs.

\paragraph{Details of downstream tasks.} For uni-modal downstream tasks, we used the AdamW optimizer \cite{loshchilov2017decoupled} with the learning rate set to 3e-6 and 3e-4 for the pre-trained model and task-specific layers, respectively. We conducted binary classification and multi-class classification for different datasets. For image-text retrieval tasks, we computed pairwise similarity to rank paired data by relevance. For zero-shot categorization we used images as queries and generated expert text prompts as in \cite{huang2021gloria}. The text prompt with the highest score would be considered a predictive positive sample. We used images as the prompt for the medical report generation task to guide text generation. We trained two auto-regressive models in image generation, e.g. Pixelsnail \cite{chen2018pixelsnail}, to model multi-modal priors. Then we sampled latent encodings and fed them into a hierarchical decoder to generate new images.

\section{Datasets}
\label{sec:data}

\paragraph{MIMIC-CXR\cite{johnson2019mimic}} This is the largest radiology dataset currently available, comprising chest X-ray images and corresponding reports from Beth Israel Deaconess Medical Center. It contains over 370,000 images from more than 65,000 patients, making it one of the most extensive collections of de-identified chest X-rays available for research. Each image is accompanied by detailed textual reports, which provide diagnostic information and contextual clinical notes. For the use of this dataset, We exclude samples that lack a "findings" or "impression" section within their clinical reports and retain only images in the frontal view. For dataset splitting, we utilize the official train-test splits provided by MIMIC-CXR. For downstream tasks, following the approach in \cite{cheng2023prior}, we sample the \textbf{MIMIC 5x200} subset and remove it from the training set to ensure robust evaluation.

\paragraph{CheXpert 5x200} The original CheXpert dataset's chest radiographs \cite{irvin2019chexpert} are multi-labeled to accommodate numerous medical observations occurring at the same time. Because our zero-shot classification relies on identifying the most comparable target, having numerous alternative labels for a target can lead to results that are inconsistent across categories. As a result, following setting in \cite{zhang2022contrastive, huang2021gloria}, we employ CheXpert's partial data to construct the CheXpert 5x200 dataset, which has 200 solely positive images for each of the CheXpert competition tasks: atelectasis, cardiomegaly, pneumonia, edema, and pleural effusion. In this dataset, each image has positive labels for only one condition.

\paragraph{RSNA Pneumonia\cite{shih2019augmenting}} The RSNA Pneumonia Detection Challenge dataset, developed by the Radiological Society of North America (RSNA), is a large, annotated collection of chest X-ray images specifically labelled for pneumonia detection. We use the stage 2 version. This dataset contains 30k frontal view chest radiographs labeled either as ”normal” or ”Pneumonia”. We sample raw 500 positives and 500 negatives for zero-shot classification. For fine-tuning, the train/valid/test split each constitutes 70\%/15\%/15\% of the dataset, following \cite{cheng2023prior}. 

\paragraph{COVIDx \cite{Wang2020}} It includes more than 30,000 CXR pictures from a global group of more than 16,600 patients. 16, 490 positive COVID-19 pictures from more than 2,800 patients are included in this collection. We make use of version 6 of this dataset. The task is to categorize each radiograph into three groups: normal, non-COVID pneumonia, and COVID-19. The data split follows \cite{wang2022multi}.

\paragraph{SIIM-ACR Pneumothorax\cite{siim-acr-pneumothorax-segmentation}} 
The SIIM-ACR Pneumothorax Segmentation Challenge is a collaborative machine-learning competition organized by the Society for Imaging Informatics in Medicine (SIIM) and the American College of Radiology (ACR). SIIM-ACR Pneumothorax contains 12954 X-ray chest images, together with image-level pneumothorax annotation and pixel-level segmentation mask if pneumothorax exists. We use them for downstream supervised classification as in \cite{cheng2023prior}.

\section{Additional visual reconstruction}
We randomly select several reconstructed visual contents and compare them with real images (see Figure \ref{figS2}), which shows that our framework could capture visual details.

\begin{figure}[]
    \centering{\includegraphics[width=\linewidth]{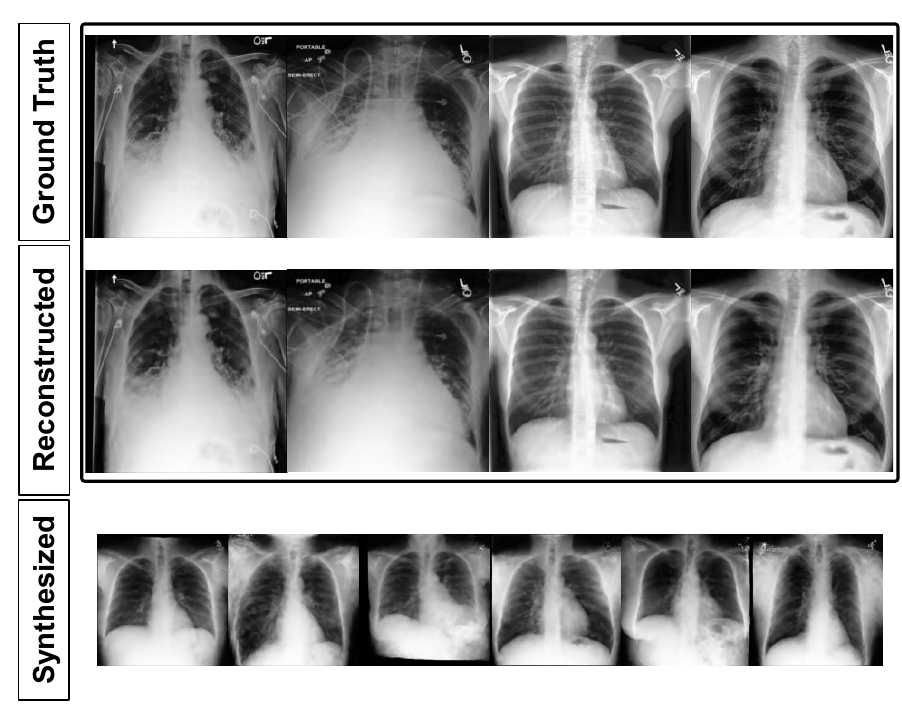}}
    \caption{Visualization of generated images. Top: real radiographs. Middle: reconstructed images corresponding to the real samples. Bottom: generated radiographs through trained Pixelsnail models and VAE decoder.}
    \label{fig3}
\end{figure}

\begin{table*}[]
\centering
\begin{tabular}{@{}c|cccc|c|ccc@{}}
\toprule
\multirow{2}{*}{ID} & \multicolumn{4}{c|}{Learning objectives} & Zero-shot cls                                                & \multicolumn{3}{c}{Fine-tuned cls}                                                                                                                                      \\ \cmidrule(l){2-9} 
                    & ITC      & ITM      & ITG      & TIG     & \begin{tabular}[c]{@{}c@{}}MIMIC 5x200 \\ (ACC)\end{tabular} & \begin{tabular}[c]{@{}c@{}}RSNA \\ (AUC)\end{tabular} & \begin{tabular}[c]{@{}c@{}}SIIM \\ (AUC)\end{tabular} & \begin{tabular}[c]{@{}c@{}}COVIDx \\ (ACC)\end{tabular} \\ \midrule
1                   &  \checkmark        &          &          &         & 41.4                                                         & 87.0                                                  & 89.4                                                  & 90.8                                                    \\
2                   &   \checkmark       &   \checkmark       &          &         & 44.3                                                         & 87.1                                                  & 89.8                                                  & 91.5                                                    \\
3                   &    \checkmark      &  \checkmark        &    \checkmark      &         & 44.8                                                         & 88.1                                                  & 92.3                                                  & 92.8                                                    \\
4                   &    \checkmark      &   \checkmark       &          &  \checkmark       & 46.2                                                         & 89.1                                                  & 92.6                                                  & 91.3                                                    \\ \midrule
5 $^\$ $                  &  \checkmark        &   \checkmark       &  \checkmark        &  \checkmark       & \textbf{50.4}                                                & \textbf{91.7}                                         & \textbf{94.8}                                         & \textbf{93.5}                                           \\
6 $^\& $                 &   \checkmark       &  \checkmark        & \checkmark         &   \checkmark      & 49.7                                                         & 87.5                                                  & 92.0                                                  & 92.5                                                    \\
7 $^\#$                  &   \checkmark       &  \checkmark        &    \checkmark      &    \checkmark     & 49.1                                                         & 90.8                                                  & 93.6                                                  & 92.0                                                    \\ \bottomrule
\end{tabular}%
\caption{Ablation studies of proposed components. \$, \& and \# represent the weight of TIG loss set to 1.0, 0.8, 1.2, respectively.}
\label{table:ablation}
\end{table*}

\section{Detailed ablation study}
\label{sec:more abl}
We further conducted comprehensive ablation studies to evaluate the performance on in-/out-of-distribution datasets and various downstream tasks in Table \ref{table:ablation}. Results highlight the effectiveness of the TIG module for enhancing cross-modal alignment. Additionally, the various loss weights were determined through empirical testing, as illustrated in Table \ref{table:ablation} ID 5, 6, 7. We examined the impact of varying the TIG loss weight while maintaining the other losses' weights constant.

%% file: main.bbl
\begin{thebibliography}{72}
\providecommand{\natexlab}[1]{#1}
\providecommand{\url}[1]{\texttt{#1}}
\expandafter\ifx\csname urlstyle\endcsname\relax
  \providecommand{\doi}[1]{doi: #1}\else
  \providecommand{\doi}{doi: \begingroup \urlstyle{rm}\Url}\fi

\bibitem[Alayrac et~al.(2022)Alayrac, Donahue, Luc, Miech, Barr, Hasson, Lenc, Mensch, Millican, Reynolds, et~al.]{alayrac2022flamingo}
Jean-Baptiste Alayrac, Jeff Donahue, Pauline Luc, Antoine Miech, Iain Barr, Yana Hasson, Karel Lenc, Arthur Mensch, Katherine Millican, Malcolm Reynolds, et~al.
\newblock Flamingo: a visual language model for few-shot learning.
\newblock \emph{Advances in neural information processing systems}, 35:\penalty0 23716--23736, 2022.

\bibitem[Bannur et~al.(2023)Bannur, Hyland, Liu, Perez-Garcia, Ilse, Castro, Boecking, Sharma, Bouzid, Thieme, et~al.]{bannur2023learning}
Shruthi Bannur, Stephanie Hyland, Qianchu Liu, Fernando Perez-Garcia, Maximilian Ilse, Daniel~C Castro, Benedikt Boecking, Harshita Sharma, Kenza Bouzid, Anja Thieme, et~al.
\newblock Learning to exploit temporal structure for biomedical vision-language processing.
\newblock In \emph{Proceedings of the IEEE/CVF Conference on Computer Vision and Pattern Recognition}, pages 15016--15027, 2023.

\bibitem[Bao et~al.(2022)Bao, Wang, Dong, Liu, Mohammed, Aggarwal, Som, Piao, and Wei]{bao2022vlmo}
Hangbo Bao, Wenhui Wang, Li Dong, Qiang Liu, Owais~Khan Mohammed, Kriti Aggarwal, Subhojit Som, Songhao Piao, and Furu Wei.
\newblock Vlmo: Unified vision-language pre-training with mixture-of-modality-experts.
\newblock \emph{Advances in Neural Information Processing Systems}, 35:\penalty0 32897--32912, 2022.

\bibitem[Boecking et~al.(2022)Boecking, Usuyama, Bannur, Castro, Schwaighofer, Hyland, Wetscherek, Naumann, Nori, Alvarez-Valle, et~al.]{boecking2022making}
Benedikt Boecking, Naoto Usuyama, Shruthi Bannur, Daniel~C Castro, Anton Schwaighofer, Stephanie Hyland, Maria Wetscherek, Tristan Naumann, Aditya Nori, Javier Alvarez-Valle, et~al.
\newblock Making the most of text semantics to improve biomedical vision--language processing.
\newblock In \emph{European conference on computer vision}, pages 1--21. Springer, 2022.

\bibitem[Brown(2020)]{brown2020language}
Tom~B Brown.
\newblock Language models are few-shot learners.
\newblock \emph{arXiv preprint arXiv:2005.14165}, 2020.

\bibitem[Chambon et~al.(2022)Chambon, Bluethgen, Delbrouck, Van~der Sluijs, Po{\l}acin, Chaves, Abraham, Purohit, Langlotz, and Chaudhari]{chambon2022roentgen}
Pierre Chambon, Christian Bluethgen, Jean-Benoit Delbrouck, Rogier Van~der Sluijs, Ma{\l}gorzata Po{\l}acin, Juan Manuel~Zambrano Chaves, Tanishq~Mathew Abraham, Shivanshu Purohit, Curtis~P Langlotz, and Akshay Chaudhari.
\newblock Roentgen: vision-language foundation model for chest x-ray generation.
\newblock \emph{arXiv preprint arXiv:2211.12737}, 2022.

\bibitem[Chen et~al.(2023{\natexlab{a}})Chen, Zhong, Wu, Luo, and Li]{chen2023contrastive}
Cheng Chen, Aoxiao Zhong, Dufan Wu, Jie Luo, and Quanzheng Li.
\newblock Contrastive masked image-text modeling for medical visual representation learning.
\newblock In \emph{International Conference on Medical Image Computing and Computer-Assisted Intervention}, pages 493--503. Springer, 2023{\natexlab{a}}.

\bibitem[Chen et~al.(2018)Chen, Mishra, Rohaninejad, and Abbeel]{chen2018pixelsnail}
Xi Chen, Nikhil Mishra, Mostafa Rohaninejad, and Pieter Abbeel.
\newblock Pixelsnail: An improved autoregressive generative model.
\newblock In \emph{International conference on machine learning}, pages 864--872. PMLR, 2018.

\bibitem[Chen et~al.(2020)Chen, Song, Chang, and Wan]{chen2020generating}
Zhihong Chen, Yan Song, Tsung-Hui Chang, and Xiang Wan.
\newblock Generating radiology reports via memory-driven transformer.
\newblock \emph{arXiv preprint arXiv:2010.16056}, 2020.

\bibitem[Chen et~al.(2023{\natexlab{b}})Chen, Diao, Wang, Li, and Wan]{chen2023towards}
Zhihong Chen, Shizhe Diao, Benyou Wang, Guanbin Li, and Xiang Wan.
\newblock Towards unifying medical vision-and-language pre-training via soft prompts.
\newblock In \emph{Proceedings of the IEEE/CVF International Conference on Computer Vision}, pages 23403--23413, 2023{\natexlab{b}}.

\bibitem[Cheng et~al.(2023)Cheng, Lin, Lyu, Huang, Luo, and Tang]{cheng2023prior}
Pujin Cheng, Li Lin, Junyan Lyu, Yijin Huang, Wenhan Luo, and Xiaoying Tang.
\newblock Prior: Prototype representation joint learning from medical images and reports.
\newblock In \emph{Proceedings of the IEEE/CVF International Conference on Computer Vision}, pages 21361--21371, 2023.

\bibitem[Denkowski and Lavie(2011)]{denkowski2011meteor}
Michael Denkowski and Alon Lavie.
\newblock Meteor 1.3: Automatic metric for reliable optimization and evaluation of machine translation systems.
\newblock In \emph{Proceedings of the sixth workshop on statistical machine translation}, pages 85--91, 2011.

\bibitem[Devlin(2018)]{devlin2018bert}
Jacob Devlin.
\newblock Bert: Pre-training of deep bidirectional transformers for language understanding.
\newblock \emph{arXiv preprint arXiv:1810.04805}, 2018.

\bibitem[Dhariwal and Nichol(2021)]{dhariwal2021diffusion}
Prafulla Dhariwal and Alexander Nichol.
\newblock Diffusion models beat gans on image synthesis.
\newblock \emph{Advances in neural information processing systems}, 34:\penalty0 8780--8794, 2021.

\bibitem[Ding et~al.(2021)Ding, Yang, Hong, Zheng, Zhou, Yin, Lin, Zou, Shao, Yang, et~al.]{ding2021cogview}
Ming Ding, Zhuoyi Yang, Wenyi Hong, Wendi Zheng, Chang Zhou, Da Yin, Junyang Lin, Xu Zou, Zhou Shao, Hongxia Yang, et~al.
\newblock Cogview: Mastering text-to-image generation via transformers.
\newblock \emph{Advances in neural information processing systems}, 34:\penalty0 19822--19835, 2021.

\bibitem[Ding et~al.(2022)Ding, Zheng, Hong, and Tang]{ding2022cogview2}
Ming Ding, Wendi Zheng, Wenyi Hong, and Jie Tang.
\newblock Cogview2: Faster and better text-to-image generation via hierarchical transformers.
\newblock \emph{Advances in Neural Information Processing Systems}, 35:\penalty0 16890--16902, 2022.

\bibitem[Dong et~al.(2019)Dong, Yang, Wang, Wei, Liu, Wang, Gao, Zhou, and Hon]{dong2019unified}
Li Dong, Nan Yang, Wenhui Wang, Furu Wei, Xiaodong Liu, Yu Wang, Jianfeng Gao, Ming Zhou, and Hsiao-Wuen Hon.
\newblock Unified language model pre-training for natural language understanding and generation.
\newblock \emph{Advances in neural information processing systems}, 32, 2019.

\bibitem[Fang et~al.(2023)Fang, Wang, Xie, Sun, Wu, Wang, Huang, Wang, and Cao]{fang2023eva}
Yuxin Fang, Wen Wang, Binhui Xie, Quan Sun, Ledell Wu, Xinggang Wang, Tiejun Huang, Xinlong Wang, and Yue Cao.
\newblock Eva: Exploring the limits of masked visual representation learning at scale.
\newblock In \emph{Proceedings of the IEEE/CVF Conference on Computer Vision and Pattern Recognition}, pages 19358--19369, 2023.

\bibitem[Gafni et~al.(2022)Gafni, Polyak, Ashual, Sheynin, Parikh, and Taigman]{gafni2022make}
Oran Gafni, Adam Polyak, Oron Ashual, Shelly Sheynin, Devi Parikh, and Yaniv Taigman.
\newblock Make-a-scene: Scene-based text-to-image generation with human priors.
\newblock In \emph{European Conference on Computer Vision}, pages 89--106. Springer, 2022.

\bibitem[Gu et~al.(2022)Gu, Chen, Bao, Wen, Zhang, Chen, Yuan, and Guo]{gu2022vector}
Shuyang Gu, Dong Chen, Jianmin Bao, Fang Wen, Bo Zhang, Dongdong Chen, Lu Yuan, and Baining Guo.
\newblock Vector quantized diffusion model for text-to-image synthesis.
\newblock In \emph{Proceedings of the IEEE/CVF conference on computer vision and pattern recognition}, pages 10696--10706, 2022.

\bibitem[Gugger et~al.(2022)Gugger, Debut, Wolf, Schmid, Mueller, Mangrulkar, Sun, and Bossan]{accelerate}
Sylvain Gugger, Lysandre Debut, Thomas Wolf, Philipp Schmid, Zachary Mueller, Sourab Mangrulkar, Marc Sun, and Benjamin Bossan.
\newblock Accelerate: Training and inference at scale made simple, efficient and adaptable.
\newblock \url{https://github.com/huggingface/accelerate}, 2022.

\bibitem[He et~al.(2016)He, Zhang, Ren, and Sun]{he2016deep}
Kaiming He, Xiangyu Zhang, Shaoqing Ren, and Jian Sun.
\newblock Deep residual learning for image recognition.
\newblock In \emph{Proceedings of the IEEE conference on computer vision and pattern recognition}, pages 770--778, 2016.

\bibitem[Heusel et~al.(2017)Heusel, Ramsauer, Unterthiner, Nessler, and Hochreiter]{heusel2017gans}
Martin Heusel, Hubert Ramsauer, Thomas Unterthiner, Bernhard Nessler, and Sepp Hochreiter.
\newblock Gans trained by a two time-scale update rule converge to a local nash equilibrium.
\newblock \emph{Advances in neural information processing systems}, 30, 2017.

\bibitem[Higgins et~al.(2017)Higgins, Matthey, Pal, Burgess, Glorot, Botvinick, Mohamed, and Lerchner]{higgins2017beta}
Irina Higgins, Loic Matthey, Arka Pal, Christopher~P Burgess, Xavier Glorot, Matthew~M Botvinick, Shakir Mohamed, and Alexander Lerchner.
\newblock beta-vae: Learning basic visual concepts with a constrained variational framework.
\newblock \emph{ICLR (Poster)}, 3, 2017.

\bibitem[Ho et~al.(2020)Ho, Jain, and Abbeel]{ho2020denoising}
Jonathan Ho, Ajay Jain, and Pieter Abbeel.
\newblock Denoising diffusion probabilistic models.
\newblock \emph{Advances in neural information processing systems}, 33:\penalty0 6840--6851, 2020.

\bibitem[Huang et~al.(2021)Huang, Shen, Lungren, and Yeung]{huang2021gloria}
Shih-Cheng Huang, Liyue Shen, Matthew~P Lungren, and Serena Yeung.
\newblock Gloria: A multimodal global-local representation learning framework for label-efficient medical image recognition.
\newblock In \emph{Proceedings of the IEEE/CVF International Conference on Computer Vision}, pages 3942--3951, 2021.

\bibitem[Huh et~al.(2023)Huh, Park, Lee, and Ye]{huh2023improving}
Jaeyoung Huh, Sangjoon Park, Jeong~Eun Lee, and Jong~Chul Ye.
\newblock Improving medical speech-to-text accuracy using vision-language pre-training models.
\newblock \emph{IEEE Journal of Biomedical and Health Informatics}, 2023.

\bibitem[Irvin et~al.(2019)Irvin, Rajpurkar, Ko, Yu, Ciurea-Ilcus, Chute, Marklund, Haghgoo, Ball, Shpanskaya, et~al.]{irvin2019chexpert}
Jeremy Irvin, Pranav Rajpurkar, Michael Ko, Yifan Yu, Silviana Ciurea-Ilcus, Chris Chute, Henrik Marklund, Behzad Haghgoo, Robyn Ball, Katie Shpanskaya, et~al.
\newblock Chexpert: A large chest radiograph dataset with uncertainty labels and expert comparison.
\newblock In \emph{Proceedings of the AAAI conference on artificial intelligence}, pages 590--597, 2019.

\bibitem[Jia et~al.(2021)Jia, Yang, Xia, Chen, Parekh, Pham, Le, Sung, Li, and Duerig]{jia2021scaling}
Chao Jia, Yinfei Yang, Ye Xia, Yi-Ting Chen, Zarana Parekh, Hieu Pham, Quoc Le, Yun-Hsuan Sung, Zhen Li, and Tom Duerig.
\newblock Scaling up visual and vision-language representation learning with noisy text supervision.
\newblock In \emph{International conference on machine learning}, pages 4904--4916. PMLR, 2021.

\bibitem[Johnson et~al.(2019)Johnson, Pollard, Greenbaum, Lungren, Deng, Peng, Lu, Mark, Berkowitz, and Horng]{johnson2019mimic}
Alistair~EW Johnson, Tom~J Pollard, Nathaniel~R Greenbaum, Matthew~P Lungren, Chih-ying Deng, Yifan Peng, Zhiyong Lu, Roger~G Mark, Seth~J Berkowitz, and Steven Horng.
\newblock Mimic-cxr-jpg, a large publicly available database of labeled chest radiographs.
\newblock \emph{arXiv preprint arXiv:1901.07042}, 2019.

\bibitem[Kang et~al.(2023)Kang, Zhu, Zhang, Park, Shechtman, Paris, and Park]{kang2023scaling}
Minguk Kang, Jun-Yan Zhu, Richard Zhang, Jaesik Park, Eli Shechtman, Sylvain Paris, and Taesung Park.
\newblock Scaling up gans for text-to-image synthesis.
\newblock In \emph{Proceedings of the IEEE/CVF Conference on Computer Vision and Pattern Recognition}, pages 10124--10134, 2023.

\bibitem[Lee et~al.(2023{\natexlab{a}})Lee, Lee, Kim, Kim, Kim, Kim, Sunwoo, and Choi]{lee2023vision}
Hyungyung Lee, Da~Young Lee, Wonjae Kim, Jin-Hwa Kim, Tackeun Kim, Jihang Kim, Leonard Sunwoo, and Edward Choi.
\newblock Vision-language generative model for view-specific chest x-ray generation.
\newblock \emph{arXiv preprint arXiv:2302.12172}, 2023{\natexlab{a}}.

\bibitem[Lee et~al.(2023{\natexlab{b}})Lee, Kim, Chang, and Ye]{lee2023llm}
Suhyeon Lee, Won~Jun Kim, Jinho Chang, and Jong~Chul Ye.
\newblock Llm-cxr: Instruction-finetuned llm for cxr image understanding and generation.
\newblock \emph{arXiv preprint arXiv:2305.11490}, 2023{\natexlab{b}}.

\bibitem[Li et~al.(2021)Li, Selvaraju, Gotmare, Joty, Xiong, and Hoi]{li2021align}
Junnan Li, Ramprasaath Selvaraju, Akhilesh Gotmare, Shafiq Joty, Caiming Xiong, and Steven Chu~Hong Hoi.
\newblock Align before fuse: Vision and language representation learning with momentum distillation.
\newblock \emph{Advances in neural information processing systems}, 34:\penalty0 9694--9705, 2021.

\bibitem[Li et~al.(2022{\natexlab{a}})Li, Li, Xiong, and Hoi]{li2022blip}
Junnan Li, Dongxu Li, Caiming Xiong, and Steven Hoi.
\newblock Blip: Bootstrapping language-image pre-training for unified vision-language understanding and generation.
\newblock In \emph{International conference on machine learning}, pages 12888--12900. PMLR, 2022{\natexlab{a}}.

\bibitem[Li et~al.(2023{\natexlab{a}})Li, Li, Savarese, and Hoi]{li2023blip}
Junnan Li, Dongxu Li, Silvio Savarese, and Steven Hoi.
\newblock Blip-2: Bootstrapping language-image pre-training with frozen image encoders and large language models.
\newblock In \emph{International conference on machine learning}, pages 19730--19742. PMLR, 2023{\natexlab{a}}.

\bibitem[Li et~al.(2023{\natexlab{b}})Li, Li, Savarese, and Hoi]{pmlr-v202-li23q}
Junnan Li, Dongxu Li, Silvio Savarese, and Steven Hoi.
\newblock {BLIP}-2: Bootstrapping language-image pre-training with frozen image encoders and large language models.
\newblock In \emph{Proceedings of the 40th International Conference on Machine Learning}, pages 19730--19742. PMLR, 2023{\natexlab{b}}.

\bibitem[Li et~al.(2023{\natexlab{c}})Li, Lin, Chen, Lin, Liang, and Chang]{li2023dynamic}
Mingjie Li, Bingqian Lin, Zicong Chen, Haokun Lin, Xiaodan Liang, and Xiaojun Chang.
\newblock Dynamic graph enhanced contrastive learning for chest x-ray report generation.
\newblock In \emph{Proceedings of the IEEE/CVF Conference on Computer Vision and Pattern Recognition}, pages 3334--3343, 2023{\natexlab{c}}.

\bibitem[Li et~al.(2022{\natexlab{b}})Li, Min, Li, and Xu]{li2022stylet2i}
Zhiheng Li, Martin~Renqiang Min, Kai Li, and Chenliang Xu.
\newblock Stylet2i: Toward compositional and high-fidelity text-to-image synthesis.
\newblock In \emph{Proceedings of the IEEE/CVF Conference on Computer Vision and Pattern Recognition}, pages 18197--18207, 2022{\natexlab{b}}.

\bibitem[Li et~al.(2023{\natexlab{d}})Li, Li, Li, Wang, Guo, Lu, Jin, Zhang, and Hong]{li2023lvit}
Zihan Li, Yunxiang Li, Qingde Li, Puyang Wang, Dazhou Guo, Le Lu, Dakai Jin, You Zhang, and Qingqi Hong.
\newblock Lvit: language meets vision transformer in medical image segmentation.
\newblock \emph{IEEE transactions on medical imaging}, 2023{\natexlab{d}}.

\bibitem[Lin et~al.(2023)Lin, Chen, Li, Lin, Xu, Zhu, Liu, Cai, Yang, Zhao, et~al.]{lin2023towards}
Bingqian Lin, Zicong Chen, Mingjie Li, Haokun Lin, Hang Xu, Yi Zhu, Jianzhuang Liu, Wenjia Cai, Lei Yang, Shen Zhao, et~al.
\newblock Towards medical artificial general intelligence via knowledge-enhanced multimodal pretraining.
\newblock \emph{arXiv preprint arXiv:2304.14204}, 2023.

\bibitem[Lin(2004)]{lin2004rouge}
Chin-Yew Lin.
\newblock Rouge: A package for automatic evaluation of summaries.
\newblock In \emph{Text summarization branches out}, pages 74--81, 2004.

\bibitem[Loshchilov(2017)]{loshchilov2017decoupled}
I Loshchilov.
\newblock Decoupled weight decay regularization.
\newblock \emph{arXiv preprint arXiv:1711.05101}, 2017.

\bibitem[M{\"u}ller et~al.(2022)M{\"u}ller, Kaissis, Zou, and Rueckert]{muller2022joint}
Philip M{\"u}ller, Georgios Kaissis, Congyu Zou, and Daniel Rueckert.
\newblock Joint learning of localized representations from medical images and reports.
\newblock In \emph{European Conference on Computer Vision}, pages 685--701. Springer, 2022.

\bibitem[Nichol et~al.(2021)Nichol, Dhariwal, Ramesh, Shyam, Mishkin, McGrew, Sutskever, and Chen]{nichol2021glide}
Alex Nichol, Prafulla Dhariwal, Aditya Ramesh, Pranav Shyam, Pamela Mishkin, Bob McGrew, Ilya Sutskever, and Mark Chen.
\newblock Glide: Towards photorealistic image generation and editing with text-guided diffusion models.
\newblock \emph{arXiv preprint arXiv:2112.10741}, 2021.

\bibitem[Papineni et~al.(2002)Papineni, Roukos, Ward, and Zhu]{papineni2002bleu}
Kishore Papineni, Salim Roukos, Todd Ward, and Wei-Jing Zhu.
\newblock Bleu: a method for automatic evaluation of machine translation.
\newblock In \emph{Proceedings of the 40th annual meeting of the Association for Computational Linguistics}, pages 311--318, 2002.

\bibitem[Radford(2018)]{radford2018improving}
Alec Radford.
\newblock Improving language understanding by generative pre-training.
\newblock 2018.

\bibitem[Radford et~al.(2021)Radford, Kim, Hallacy, Ramesh, Goh, Agarwal, Sastry, Askell, Mishkin, Clark, et~al.]{radford2021learning}
Alec Radford, Jong~Wook Kim, Chris Hallacy, Aditya Ramesh, Gabriel Goh, Sandhini Agarwal, Girish Sastry, Amanda Askell, Pamela Mishkin, Jack Clark, et~al.
\newblock Learning transferable visual models from natural language supervision.
\newblock In \emph{International conference on machine learning}, pages 8748--8763. PMLR, 2021.

\bibitem[Ramesh et~al.(2021)Ramesh, Pavlov, Goh, Gray, Voss, Radford, Chen, and Sutskever]{ramesh2021zero}
Aditya Ramesh, Mikhail Pavlov, Gabriel Goh, Scott Gray, Chelsea Voss, Alec Radford, Mark Chen, and Ilya Sutskever.
\newblock Zero-shot text-to-image generation.
\newblock In \emph{International conference on machine learning}, pages 8821--8831. Pmlr, 2021.

\bibitem[Razavi et~al.(2019)Razavi, Van~den Oord, and Vinyals]{razavi2019generating}
Ali Razavi, Aaron Van~den Oord, and Oriol Vinyals.
\newblock Generating diverse high-fidelity images with vq-vae-2.
\newblock \emph{Advances in neural information processing systems}, 32, 2019.

\bibitem[Reed et~al.(2016)Reed, Akata, Yan, Logeswaran, Schiele, and Lee]{reed2016generative}
Scott Reed, Zeynep Akata, Xinchen Yan, Lajanugen Logeswaran, Bernt Schiele, and Honglak Lee.
\newblock Generative adversarial text to image synthesis.
\newblock In \emph{International conference on machine learning}, pages 1060--1069. PMLR, 2016.

\bibitem[Reyes et~al.(2020)Reyes, Meier, Pereira, Silva, Dahlweid, Tengg-Kobligk, Summers, and Wiest]{reyes2020interpretability}
Mauricio Reyes, Raphael Meier, S{\'e}rgio Pereira, Carlos~A Silva, Fried-Michael Dahlweid, Hendrik~von Tengg-Kobligk, Ronald~M Summers, and Roland Wiest.
\newblock On the interpretability of artificial intelligence in radiology: challenges and opportunities.
\newblock \emph{Radiology: artificial intelligence}, 2\penalty0 (3):\penalty0 e190043, 2020.

\bibitem[Shih et~al.(2019)Shih, Wu, Halabi, Kohli, Prevedello, Cook, Sharma, Amorosa, Arteaga, Galperin-Aizenberg, et~al.]{shih2019augmenting}
George Shih, Carol~C Wu, Safwan~S Halabi, Marc~D Kohli, Luciano~M Prevedello, Tessa~S Cook, Arjun Sharma, Judith~K Amorosa, Veronica Arteaga, Maya Galperin-Aizenberg, et~al.
\newblock Augmenting the national institutes of health chest radiograph dataset with expert annotations of possible pneumonia.
\newblock \emph{Radiology: Artificial Intelligence}, 1\penalty0 (1):\penalty0 e180041, 2019.

\bibitem[Song et~al.(2020)Song, Meng, and Ermon]{song2020denoising}
Jiaming Song, Chenlin Meng, and Stefano Ermon.
\newblock Denoising diffusion implicit models.
\newblock \emph{arXiv preprint arXiv:2010.02502}, 2020.

\bibitem[Su et~al.(2019)Su, Zhu, Cao, Li, Lu, Wei, and Dai]{su2019vl}
Weijie Su, Xizhou Zhu, Yue Cao, Bin Li, Lewei Lu, Furu Wei, and Jifeng Dai.
\newblock Vl-bert: Pre-training of generic visual-linguistic representations.
\newblock \emph{arXiv preprint arXiv:1908.08530}, 2019.

\bibitem[Van Den~Oord et~al.(2017)Van Den~Oord, Vinyals, et~al.]{van2017neural}
Aaron Van Den~Oord, Oriol Vinyals, et~al.
\newblock Neural discrete representation learning.
\newblock \emph{Advances in neural information processing systems}, 30, 2017.

\bibitem[Vaswani(2017)]{vaswani2017attention}
A Vaswani.
\newblock Attention is all you need.
\newblock \emph{Advances in Neural Information Processing Systems}, 2017.

\bibitem[Wang et~al.(2022{\natexlab{a}})Wang, Zhou, Wang, Vardhanabhuti, and Yu]{wang2022multi}
Fuying Wang, Yuyin Zhou, Shujun Wang, Varut Vardhanabhuti, and Lequan Yu.
\newblock Multi-granularity cross-modal alignment for generalized medical visual representation learning.
\newblock \emph{Advances in Neural Information Processing Systems}, 35:\penalty0 33536--33549, 2022{\natexlab{a}}.

\bibitem[Wang et~al.(2020)Wang, Lin, and Wong]{Wang2020}
Linda Wang, Zhong~Qiu Lin, and Alexander Wong.
\newblock Covid-net: a tailored deep convolutional neural network design for detection of covid-19 cases from chest x-ray images.
\newblock \emph{Scientific Reports}, 10\penalty0 (1):\penalty0 19549, 2020.

\bibitem[Wang et~al.(2022{\natexlab{b}})Wang, Bao, Dong, Bjorck, Peng, Liu, Aggarwal, Mohammed, Singhal, Som, et~al.]{wang2022image}
Wenhui Wang, Hangbo Bao, Li Dong, Johan Bjorck, Zhiliang Peng, Qiang Liu, Kriti Aggarwal, Owais~Khan Mohammed, Saksham Singhal, Subhojit Som, et~al.
\newblock Image as a foreign language: Beit pretraining for all vision and vision-language tasks.
\newblock \emph{arXiv preprint arXiv:2208.10442}, 2022{\natexlab{b}}.

\bibitem[Wang and Liu(2019)]{wang2019translating}
Zelun Wang and Jyh-Charn Liu.
\newblock Translating math formula images to latex sequences using deep neural networks with sequence-level training, 2019.

\bibitem[Wang et~al.(2022{\natexlab{c}})Wang, Wu, Agarwal, and Sun]{wang2022medclip}
Zifeng Wang, Zhenbang Wu, Dinesh Agarwal, and Jimeng Sun.
\newblock Medclip: Contrastive learning from unpaired medical images and text.
\newblock In \emph{2022 Conference on Empirical Methods in Natural Language Processing, EMNLP 2022}, 2022{\natexlab{c}}.

\bibitem[Wu et~al.(2023)Wu, Zhang, Zhang, Wang, and Xie]{wu2023medklip}
Chaoyi Wu, Xiaoman Zhang, Ya Zhang, Yanfeng Wang, and Weidi Xie.
\newblock Medklip: Medical knowledge enhanced language-image pre-training for x-ray diagnosis.
\newblock In \emph{Proceedings of the IEEE/CVF International Conference on Computer Vision}, pages 21372--21383, 2023.

\bibitem[You et~al.(2023)You, Gu, Ham, Park, Kim, Hong, Baek, and Roh]{you2023cxr}
Kihyun You, Jawook Gu, Jiyeon Ham, Beomhee Park, Jiho Kim, Eun~K Hong, Woonhyuk Baek, and Byungseok Roh.
\newblock Cxr-clip: Toward large scale chest x-ray language-image pre-training.
\newblock In \emph{International Conference on Medical Image Computing and Computer-Assisted Intervention}, pages 101--111. Springer, 2023.

\bibitem[Yu et~al.(2022)Yu, Xu, Koh, Luong, Baid, Wang, Vasudevan, Ku, Yang, Ayan, et~al.]{yu2022scaling}
Jiahui Yu, Yuanzhong Xu, Jing~Yu Koh, Thang Luong, Gunjan Baid, Zirui Wang, Vijay Vasudevan, Alexander Ku, Yinfei Yang, Burcu~Karagol Ayan, et~al.
\newblock Scaling autoregressive models for content-rich text-to-image generation.
\newblock \emph{arXiv preprint arXiv:2206.10789}, 2\penalty0 (3):\penalty0 5, 2022.

\bibitem[Zawacki et~al.(2019)Zawacki, Wu, Shih, Elliott, Fomitchev, Hussain, ParasLakhani, Culliton, and Bao]{siim-acr-pneumothorax-segmentation}
Anna Zawacki, Carol Wu, George Shih, Julia Elliott, Mikhail Fomitchev, Mohannad Hussain, ParasLakhani, Phil Culliton, and Shunxing Bao.
\newblock Siim-acr pneumothorax segmentation.
\newblock \url{https://kaggle.com/competitions/siim-acr-pneumothorax-segmentation}, 2019.
\newblock Kaggle.

\bibitem[Zhan et~al.(2024)Zhan, Lin, Wang, Wang, and Wu]{zhan2024medm2g}
Chenlu Zhan, Yu Lin, Gaoang Wang, Hongwei Wang, and Jian Wu.
\newblock Medm2g: Unifying medical multi-modal generation via cross-guided diffusion with visual invariant.
\newblock In \emph{Proceedings of the IEEE/CVF Conference on Computer Vision and Pattern Recognition}, pages 11502--11512, 2024.

\bibitem[Zhang et~al.(2021)Zhang, Koh, Baldridge, Lee, and Yang]{zhang2021cross}
Han Zhang, Jing~Yu Koh, Jason Baldridge, Honglak Lee, and Yinfei Yang.
\newblock Cross-modal contrastive learning for text-to-image generation.
\newblock In \emph{Proceedings of the IEEE/CVF conference on computer vision and pattern recognition}, pages 833--842, 2021.

\bibitem[Zhang et~al.(2022)Zhang, Jiang, Miura, Manning, and Langlotz]{zhang2022contrastive}
Yuhao Zhang, Hang Jiang, Yasuhide Miura, Christopher~D Manning, and Curtis~P Langlotz.
\newblock Contrastive learning of medical visual representations from paired images and text.
\newblock In \emph{Machine Learning for Healthcare Conference}, pages 2--25. PMLR, 2022.

\bibitem[Zhong et~al.(2023)Zhong, Xu, Liang, Chen, and Wu]{zhong2023ariadne}
Yi Zhong, Mengqiu Xu, Kongming Liang, Kaixin Chen, and Ming Wu.
\newblock Ariadne’s thread: Using text prompts to improve segmentation of infected areas from chest x-ray images.
\newblock In \emph{International Conference on Medical Image Computing and Computer-Assisted Intervention}, pages 724--733. Springer, 2023.

\bibitem[Zhou et~al.(2022{\natexlab{a}})Zhou, Chen, Zhang, Luo, Wang, and Yu]{zhou2022generalized}
Hong-Yu Zhou, Xiaoyu Chen, Yinghao Zhang, Ruibang Luo, Liansheng Wang, and Yizhou Yu.
\newblock Generalized radiograph representation learning via cross-supervision between images and free-text radiology reports.
\newblock \emph{Nature Machine Intelligence}, 4\penalty0 (1):\penalty0 32--40, 2022{\natexlab{a}}.

\bibitem[Zhou et~al.(2022{\natexlab{b}})Zhou, Lian, Wang, and Yu]{zhou2022advancing}
Hong-Yu Zhou, Chenyu Lian, Liansheng Wang, and Yizhou Yu.
\newblock Advancing radiograph representation learning with masked record modeling.
\newblock In \emph{The Eleventh International Conference on Learning Representations}, 2022{\natexlab{b}}.

\end{thebibliography}
